\documentclass[preprint,12pt]{elsarticle}
\usepackage[margin=2.5cm]{geometry}
\usepackage{amssymb}
\usepackage{amsmath}

\usepackage{graphicx}%
\usepackage{subcaption}
\usepackage{booktabs}
\usepackage{colortbl}
\usepackage[table,xcdraw]{xcolor}
\usepackage{float} 
\usepackage{adjustbox}
\usepackage{comment}
\usepackage{longtable}
\usepackage{rotating}
\usepackage{url}

\renewcommand{\arraystretch}{1.2} 

\begin{document}

\begin{frontmatter}



\title{SpreadPy: A Python tool for modelling spreading activation and superdiffusion in cognitive multiplex networks} 

\author[aff1,aff4]{Salvatore Citraro}
\author[aff2,aff4]{Edith Haim}
\author[aff2]{Alessandra Carini}
\author[aff3]{Cynthia S. Q. Siew}
\author[aff1]{Giulio Rossetti}
\author[aff2]{Massimo Stella}

\affiliation[aff1]{
  organization={Institute of Information Science and Technologies “A. Faedo” (ISTI), National Research Council (CNR)},
  city={Pisa},
  postcode={56127},
  country={Italy}
}
\affiliation[aff2]{
  organization={CogNosco Lab, Department of Psychology and Cognitive Science, University of Trento},
  city={Rovereto},
  postcode={38068},
  country={Italy}
}
\affiliation[aff3]{
  organization={Department of Psychology, National University of Singapore},
  city={Singapore},
  postcode={117570},
  country={Singapore\\}
}
\affiliation[aff4]{
  country={These authors contributed equally\\}
}

\begin{abstract}
We introduce SpreadPy as a Python library for simulating spreading activation in cognitive single-layer and multiplex networks. Our tool is designed to perform numerical simulations testing structure-function relationships in cognitive processes. By comparing simulation results with grounded theories in knowledge modelling, SpreadPy enables systematic investigations of how activation dynamics reflect cognitive, psychological and clinical phenomena. We demonstrate the library's utility through three case studies: (1) Spreading activation on associative knowledge networks distinguishes students with high versus low math anxiety, revealing anxiety-related structural differences in conceptual organization; (2) Simulations of a creativity task show that activation trajectories vary with task difficulty, exposing how cognitive load modulates lexical access; (3) In individuals with aphasia, simulated activation patterns on lexical networks correlate with empirical error types (semantic vs. phonological) during picture-naming tasks, linking network structure to clinical impairments. SpreadPy's flexible framework allows researchers to model these processes using empirically derived or theoretical networks, providing mechanistic insights into individual differences and cognitive impairments. The library is openly available, supporting reproducible research in psychology, neuroscience, and education research.
\end{abstract}





\end{frontmatter}



\section{Introduction}

Spreading activation is a network-based model describing how activation propagates through interconnected entries during information search. The original model, introduced in cognitive science \cite{collins1975spreading, anderson1983spreading}, considers the search process as the spread of an activation signal propagating on a cognitive or information network. In cognitive networks, nodes represent concepts and/or words while links indicate semantic, phonological, syntactic and other types of conceptual relationships \cite{siew2019spreadr}. In information networks, nodes could represent websites, posts or documents and they could be linked according to their similarities \cite{crestani1997application}. In spreading activation, the activation signal originates from one or multiple initial nodes and spreads outward, diminishing with each step according to predefined parameters such as decay or thresholds. The final pattern of activation identifies structurally relevant nodes, related to the origin \cite{collins1975spreading}.

In cognitive science, spreading activation models human cognitive processes \cite{anderson1983spreading,collins1969retrieval,collins1992phonological}, including lexical access and memory retrieval. For instance, when individuals read or hear a word, activation may be spreading along a networked cognitive repository of knowledge, a system known as mental lexicon and investigated across psychology, linguistics and artificial intelligence \cite{aitchison2012words}. Spreading activation can explain phenomena such as semantic priming, where exposure to a word (e.g., "doctor") speeds up the recognition of a related word (e.g., "nurse") \cite{mcrae2005semantic,aitchison2012words}. Activation spreading also accounts for lexical retrieval errors, such as tip-of-the-tongue states, where partial activation of a target word leads to the retrieval of semantically or phonologically similar words, thereby hindering the full activation and retrieval of the target word \cite{burke1991tip}. The modelling framework of activation spread has profoundly influenced our understanding of semantic memory, lexical access, and associative reasoning \cite{aitchison2012words,vitevitch2008can,mcclelland1981interactive}. However, previous implementations of spreading activation have been limited to computational frameworks in single-layer networks and via R only \cite{siew2019spreadr}. 

In computer science, spreading activation algorithms have been mainly applied to model semantic search and information retrieval tasks. They operate by activating nodes representing search queries and propagating this activation through networks of interconnected documents, concepts, or data entities. Highly activated nodes correspond to documents or entities semantically related to the original query \cite{crestani1997application}. For example, Crestani \cite{crestani1997application} demonstrated how spreading activation enhances search efficiency by retrieving documents contextually similar to user queries based on their semantic relationships. Similar works proposed a spreading activation-based navigation and information-foraging method to support users in exploring large document collections \cite{pirolli1996scatter} or the World Wide Web \cite{crestani2000searching,rocha2004hybrid}. Spreading activation was even featured in digital library systems to improve retrieval accuracy and contextual relevance among library entries, notes and digital books \cite{vsevcech2014user}. Wherever multiple types of nodes or links can exist, such as in the digital library example, spreading activation can be extended beyond single-layer structures. Multiplex networks, comprising multiple interconnected layers of information, provide a powerful framework for modelling spreading activation in these complex relationships \cite{stella2024cognitive}.

To bridge cognitive insights with novel computational applications across multiplex networks, we here present SpreadPy, a Python library designed to implement spreading activation models on both single-layer and multiplex network structures. SpreadPy facilitates the modelling of complex cognitive processes and enhances algorithmic performance in semantic search tasks. This makes it a unified framework that is valuable for both cognitive science research and artificial intelligence applications.

\subsection{Spreading activation in knowledge modelling}

Despite being introduced at the fringe of cognitive science and cognitive psychology, spreading activation has been extensively adopted and further developed in computer science. The works by Crestani and colleagues \cite{crestani1997application,crestani2000searching} investigated spreading activation techniques in relation to information retrieval. Crestani \cite{crestani1997application} introduced spreading activation as a computational mechanism to enhance semantic relevance and retrieval precision in document databases, highlighting its ability to naturally model associative thinking. Extending this concept, Crestani and Lee \cite{crestani2000searching} developed a constrained spreading activation algorithm tailored for web searching. They significantly improved query specificity by limiting activation propagation through selective semantic paths, thereby balancing retrieval comprehensiveness and efficiency.

Further developments expanded spreading activation techniques to more structured semantic domains. Rocha and colleagues \cite{rocha2004hybrid} presented a hybrid search approach integrating traditional keyword matching and spreading activation within semantic web frameworks. Their method improved retrieval performance by activating relevant semantic pathways derived from structured metadata, demonstrating the advantage of combining symbolic reasoning with spreading activation for handling complex information structures on the semantic web. Similarly, Vsevcech and colleagues \cite{vsevcech2014user} leveraged spreading activation in combination with user-generated annotations to enhance the contextual relevance of document searches within digital libraries. Consequently, they reinforced the adaptability of spreading activation to leverage both formal semantics and user-driven metadata.

In contrast to structured semantic approaches, Pirolli and colleagues \cite{pirolli1996scatter} applied spreading activation within user-centric interfaces for navigating large text collections. Their "Scatter/Gather" browsing approach used spreading activation to dynamically cluster (gather) and present specific document groups (scatter) based on topical relevance, supporting users' natural cognitive strategies of incremental exploration and information foraging. This approach differed fundamentally from structured semantic methods by emphasizing user interactions and cognitive relevance directly rather than explicit semantic structures. 

Collectively, the above studies illustrate the versatility and efficacy of spreading activation, underscoring its applicability across diverse information retrieval contexts, ranging from structured semantic databases to user-driven interactive environments. While our introduced tool SpreadPy can account for a wide variety of knowledge modelling approaches, we here focus on testing it within the field of cognitive network science, a novel research area adopting network science to investigate cognitive phenomena \cite{stella2022network,castro2020contributions,siew2019cognitive}.

\subsection{Spreading activation in cognitive network science}

The spreading activation theory implemented in SpreadPy is based on the seminal work of Collins and Loftus \cite{collins1975spreading}, whose spreading activation theory of semantic processing is built directly on Quillian's earlier model of memory search \cite{quillian1967word}. Quillian's theory introduced the idea that concepts are represented as nodes in a network and that activation spreads along associative links between them. Once a concept is stimulated through reading, hearing or mental activation (i.e. thinking about it), its activation spreads to related concepts along these links. Activation weakens with distance and time, and a threshold of cumulated activation must be reached for a concept to become accessible. 

In contrast to Quillian's model \cite{quillian1967word}, Collins and Loftus’ spreading activation model \cite{collins1975spreading} was not just intended as a theoretical explanation but as a computational framework to simulate cognitive search processes \cite{collins1975spreading}. It shared key goals with early artificial intelligence since it aimed to understand knowledge representation and memory search in humans as well as simulate human-like behaviour in computers. The model has a lasting influence on cognitive psychology, where it provided explanations for semantic priming \cite{de1983range,mcrae2005semantic}, lexical decision-making \cite{balota1986depth}, and memory retrieval phenomena \cite{saunders2006can}, among others. It offered a more flexible alternative to previous models of memory and inspired empirical research into associative strength, inhibition mechanisms, reaction times, and the structure of semantic networks \cite{saunders2006can,siew2019spreadr}.

One key advance in this area was the recent work by Siew \cite{siew2019spreadr}, which introduced the R package \textit{spreadr} to transform theoretical considerations into computational, psychological simulations. Siew's work was the first implementation of spreading activation into an R environment, able to explain and reproduce psychological results from real-world data \cite{siew2019spreadr}. While Siew's \textit{spreadr} package constitutes a significant advancement in network research, this tool operated only on single-layer networks, leaving a gap in the field for multiplex representations. 
Such advanced network models are relevant since cognitive processes, especially those involved in language and memory, have been shown to be influenced by multiple types of associations interacting with each other, such as semantic and phonological \cite{castro2020contributions}. For instance, during the lexical decision task, which heavily relies on semantic information, studies have found phonological priming effects that aid in lexical retrieval \cite{coltheart1979phonological,collins1992phonological}. 

\begin{figure}[t!]
\centering
\includegraphics[scale=0.38]{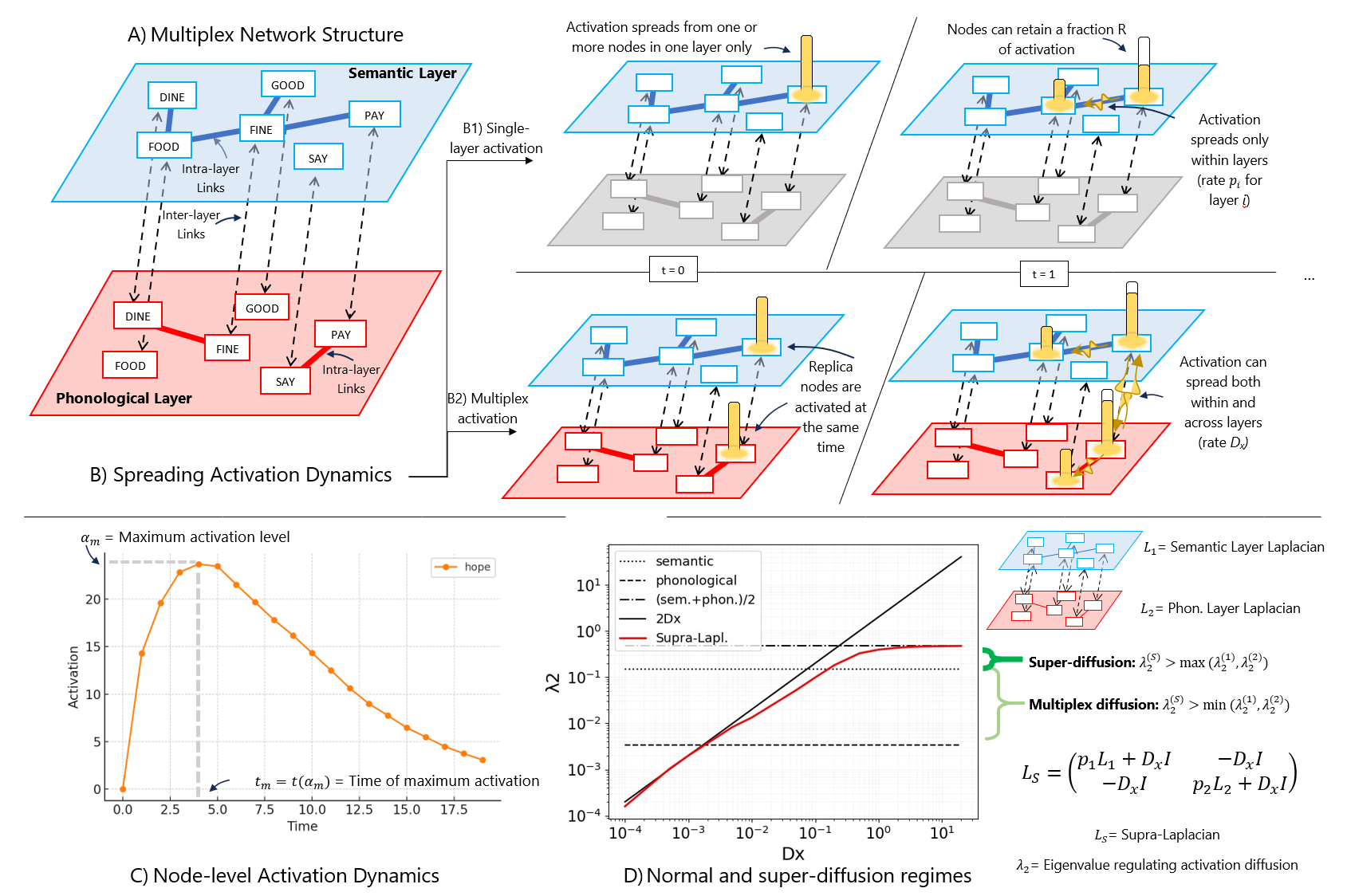}
\caption{Infographics for the functioning of SpreadPy over a cognitive multiplex network. A) A cognitive multiplex network can be a multiplex network with a semantic layer and a phonological layer, where nodes are replicated (replica nodes) across layers. In a given layer, replica nodes are connected by intra-layer connections, e.g. phonological similarities or free associations ("good"-"fine" on semantic layer). Replicas of the same node/concept are connected by inter-layer connections ("good" on semantic layer -"good" on phonological layer). B) As input, SpreadPy uses individual layers or multiplex networks to perform two types of activation spreading: (i) single-layer or (ii) multiplex. C) Example of activation for "hope". At the node level, activation arrives as a function of time, peaking at time $t_m$ (maximum activation time) and reaching at most activation level of $\alpha_m$. D) A multiplex structure can accelerate diffusion - activation on the multiplex network spreads faster than on its slowest layer. It can also cause a superdiffusion process - activation on the multiplex network spreads faster than on its fastest layer.}
\label{fig:multiplex}
\end{figure}

To capture this complexity, and fully concretize Collins and Loftus' original idea of featuring more networks linked together \cite{collins1975spreading}, we use cognitive multiplex networks \cite{stella2017multiplex,stella2018multiplex,stella2024cognitive}. These network models can incorporate multiple aspects of the mental lexicon (e.g., semantic associations and phonological similarities) at the same time. Despite their theoretical promise, computational tools for simulating spreading activation within network-based cognitive models remain underdeveloped. SpreadPy extends this approach to capture the multiplex nature of cognitive systems in which concepts are linked through diverse relations (e.g., semantic, phonological, or contextual). Figure \ref{fig:multiplex} depicts this multiplex structure. Each node represents a word, existing as replicas simultaneously in both semantic and phonological layers, connected by inter-layer links. Within each layer, nodes are connected based on layer-specific similarities: meaning-related associations in the semantic layer, and phonological similarities in the phonological layer (see Figure \ref{fig:multiplex}A). This multiplex design allows activation to spread not only within a single type of relationship (within only one layer), but also across different association types (across multiple layers), reflecting the interactive, complex nature of cognitive processes \cite{castro2020contributions}. Importantly, access to simulations based on cognitive multiplex networks also provides access to exploring the phenomenon of superdiffusion \cite{gomez2013diffusion} within cognitive science (see further Section 1.4).

\subsection{Spreading activation with SpreadPy}

Our SpreadPy approach can model both single layer activation (Figure \ref{fig:multiplex} B1) and multiplex activation (Figure \ref{fig:multiplex} B2). In single-layer activation (Figure \ref{fig:multiplex} B1), a node spreads the activation it has received only within one layer. For example, receiving a payment notification on the smartphone may activate the concept "pay" on the semantic layer. The activated node "pay" retains a fraction of its original activation (e.g. 50 percent of activation are retained) and spreads the remaining activation within the semantic layer to the neighbouring concept "fine". At timestep $t=1$, both nodes now hold 50 percent of the initial activation. From there, activation will continue to spread within the semantic layer to adjacent nodes. 
In contrast, with a multiplex model (Figure \ref{fig:multiplex} B2), we can model all replicas of a node on different layers to receive activation at the same time. For instance, while receiving a payment notification on the phone, one may read the word "pay" in the pop-up, activating the replica nodes of "pay" both on the semantic and phonological layers simultaneously. From both replicas of "pay", activation starts spreading along intra-layer links to the adjacent nodes "fine" on the semantic layer and "say" on the phonological layer. Additionally, activation can now also spread across layers, such that "fine" on the semantic layer also activates its replica on the phonological layer along an inter-layer link.

In our simulations, activation is seeded at cue words and spreads through the network over discrete time steps. We measure two key metrics: the maximum activation level a node accumulates and the time it takes to reach this peak. Figure \ref{fig:multiplex}C shows an example of how we trace the activation level of the target word "hope" over time, peaking at time $t_m$ with its maximum activation $\alpha_m$. 

Figure \ref{fig:multiplex}D further illustrates diffusive processes and the differences between normal and superdiffusion regimes in our simulations. In complex networks, diffusive processes are relative to the spread of information, signals, social norms, diseases and other attributes across nodes in a network \cite{haim2023cognitive,newman2018networks}. When activation spreads along linked nodes, network topology drives the diffusion dynamics, i.e. how quickly a signal spreads across nodes in the network. Thus, diffusion dynamics can be modelled in terms of adjacency/links between nodes, encapsulated within a binary adjacency matrix $A$, and in terms of network degree of individual nodes, encapsulated within a diagonal matrix $D$ with degrees on its main diagonal \cite{haim2023cognitive,newman2018networks}. Node degrees can indicate how much signal in total can accumulate over nodes, while individual links can indicate potential incoming or escaping routes for diffusive signals. One can encapsulate both these elements in a matrix called Laplacian, $L=D-A$, which can be normalized in several ways. The eigenspectrum of the Laplacian $L$ contains insights about the diffusion dynamics. In particular, the second smallest eigenvalue $\lambda_2$ encodes information about the characteristic diffusion time $\tau$, i.e. the time it takes for diffusion to spread along a specific region of a complex network. Through series expansion, one can check that in simple diffusion processes on complex networks, $\lambda_2 \propto 1/\tau$, i.e. the larger the $\lambda_2$ eigenvalue of the Laplacian, the shorter the time required for the diffusive signal to reach a whole region of the network. In other words, higher $\lambda_2$ means faster diffusion \cite{haim2023cognitive,koponen2021systemic}. In multiplex networks, one can build a supra-Laplacian matrix, see also Figure \ref{fig:multiplex}D, and compare the eigenvalue $\lambda_2^{(s)}$ of the whole multiplex network against the eigenvalues $\lambda_2^{(1)}$ and $\lambda_2^{(2)}$ of the individual layers. 

\subsection{Superdiffusion and its relevance for knowledge modelling}

Multiplex diffusion occurs when the diffusion on the whole multiplex network can efficiently exploit inter-layer jumps to spread at a rate faster than on the slowest layer. Inter-layer jumps can occur with a parameter $D_x$, whose magnitude regulates the rate of diffusion among replicas of the same physical node compared to the rates, $p_1$ and $p_2$, of diffusion between different replica nodes in the same layer, either layer 1 or layer 2. In other words, multiplex diffusion occurs when $\lambda_2^{(s)}$ is greater than the smallest respective eigenvalue for the Laplacians of the individual layers. In multiplex diffusion, multiplexity is beneficial because it provides shortcuts facilitating the spread of signals even among nodes being remote in individual layers. The combination of two layers together thus speeds up diffusion compared to what would happen in the slowest layer \cite{gomez2013diffusion,koponen2021systemic}. However, in this regime, diffusion over the whole multiplex remains slower than in the fastest layer. In formulas,  $\lambda_2^{(s)}$ (highlighted in red in Fig. \ref{fig:multiplex}C) would be higher than the $\text{min}(\lambda_2^{(1)},\lambda_2^{(2)})$ but lower than the $\text{max}(\lambda_2^{(1)},\lambda_2^{(2)})$.

Superdiffusion occurs when the two layers are so strongly entwined that diffusion on the multiplex network is faster even than diffusion in the faster layer, if considered in isolation. In formulays, $\lambda_2^{(s)}$ would be higher than $\text{max}(\lambda_2^{(1)},\lambda_2^{(2)})$ but still lower than the upper bound given by the superposition of Laplacians (see Fig. \ref{fig:multiplex}C). This superposition posits an upper theoretical bound to the speed of diffusion, which is the fastest when all edges in the multiplex networks are fully available at the same time, i.e. when the coupling of layer $D_x\rightarrow +\infty$ \cite{koponen2021systemic}.

The phenomenon of superdiffusion offers a novel lens for understanding cognitive processes. In cognitive systems, representations are often distributed across multiple modalities (e.g. semantic, phonological, and contextual layers), each contributing uniquely to information processing. Superdiffusion arises when interactions between these layers create synergistic effects, enabling rapid and coordinated activation across the network. This contrasts with traditional single-layer models, where activation is confined to a single modality, potentially underestimating the speed and flexibility of cognitive operations \cite{haim2023cognitive,koponen2021systemic}.  

When modelling mental search, superdiffusion acts in terms of mixing the topological information (e.g., cycles, shortest paths, neighbourhoods) of the considered network layers. Superdiffusion would thus be relative to a highly interactive set of cognitive processes taking place on all considered network layers, e.g., phonological associations and semantic associations both contributing to producing mixed errors during picture naming by people affected by aphasia \cite{dell1997lexical}. By modelling single-layer diffusion, multiplex and superdiffusion, SpreadPy enables researchers to quantify the impact of both network structure and spreading dynamics over a wide variety of phenomena. We further explore and test this in our case studies.

\subsection{Manuscript aims and outline}

This work introduces SpreadPy, a Python library that simulates spreading activation in single-layer and multiplex cognitive networks, offering researchers a flexible platform to test hypotheses about structure-function relationships in the mental lexicon. By integrating principles from cognitive psychology and network science, SpreadPy enables the exploration of how cognitive processes are modulated by network topology, such as clustering, centrality, and layer coupling.

We test and demonstrate the utility of SpreadPy through three case studies: (1) examining the impact of math anxiety on associative knowledge networks, (2) simulating activation dynamics to differentiate item difficulty in a creativity task, and (3) modelling lexical retrieval errors in aphasia. These applications highlight how SpreadPy can uncover mechanistic insights into cognitive phenomena and clinical conditions, advancing both theoretical and applied research in cognitive data science.  

\section{Materials and Methods}

This Section outlines the implementation of diffusion and superdiffusion regimes in SpreadPy, and the common network data used for simulations in this manuscript across the three case studies.

\noindent \textbf{Single-layer Diffusion}. The base model operates on a simple, undirected graph, $\mathcal{G} = (V, E)$, where $V$ is the set of nodes, and $E \subseteq V \times V$ denotes the set of edges as unordered pairs of nodes.
Each node holds an amount of activation energy that can be transferred to its neighbours over time.
At each time step $t$, an active node $u \in V$ with non-zero energy $e_{u,t}$ retains a fraction $R$ of its energy and distributes the remaining portion uniformly among its neighbours.
The energy transfer from node $u$ to a neighbour $v \in \Gamma_u$, being $\Gamma_u$ the set of $u$'s neighbours, is given by the following equation:

\begin{equation}
\varphi(u, v) = \frac{e_{u,t} \cdot (1 - R)}{\left| \Gamma_u \right|},
\end{equation}

where $\left| \Gamma_u \right|$ is the number of $u$'s neighbours, and $R$ is the retention factor controlling the energy kept within $u$.
\\ \ \\
\noindent \textbf{Multiplex Diffusion}. The network is extended to a multiplex graph, $\mathcal{M} = (V, E, L)$, where $L$ is the set of layers.
Each node $u \in V$ can exist as a replica $u^{[\ell]}$ in one or more layers, indexed by $\ell \in L$.
Edges can be within the same layer (intra-) or between replica nodes across layers (inter-).
A coupling parameter $D_x > 0$ balances intra- and inter-layer diffusion, with probabilities $p_{\parallel} = \frac{1}{1 + D_x}$ and $p_{\perp} = \frac{D_x}{1 + D_x}$, respectively.

Intra-layer spreading mirrors the single-layer case, scaled by $p_{\parallel}$.
The energy transfer from node $u^{[\ell]}$ to a neighbour $v^{[\ell]} \in \Gamma_u^{[\ell]}$, being $\Gamma_u^{[\ell]}$ the set of $u$'s neighbours on layer $l$, is given by the equation:

\begin{equation}
\varphi(u^{[\ell]}, v^{[\ell]}) = \frac{e_{u^{[\ell]}, t} \cdot (1 - R)}{|\Gamma_u^{[\ell]}|} \cdot p_{\parallel},
\end{equation}

where $|\Gamma_u^{[\ell]}|$ is the number of $u$'s neighbours on layer $l$.

Inter-layer spreading allows a node to transfer activation energy to its replica in another layer, after which the energy continues to spread among the neighbours in the new layer.
This process can be described by dividing it into two parts: replica energy transfer and neighbours inter-layer spreading. 
Energy transfer from node $u^{[\ell]}$ to its replica $u^{[\ell']}$, with $l' \neq l$, is modelled as follows:

\begin{equation}
\varphi(u^{[\ell]}, u^{[\ell']}) = e_{u^{[\ell]}, t} \cdot (1 - R) \cdot p_{\perp}.
\end{equation}

Neighbours inter-layer spreading describes activation passing from a node on one layer to its replica on another layer, and then continues spreading to that replica's neighbour within the second layer. Such neighbours inter-layer spreading occurs in $l'$, where the replica energy is transferred among $u$'s neighbours in $\Gamma_u^{[\ell']}$ as follows:

\begin{equation}
\varphi(u^{[\ell']}, v^{[\ell']}) = \frac{\varphi(u^{[\ell]}, u^{[\ell']})}{|\Gamma_u^{[\ell']}|},
\end{equation}

where $|\Gamma_u^{[\ell']}|$ is the number of neighbours of $u$ on layer $l'$.

This formulation ensures that the total transferred energy (after retention) is divided between intra- and inter-layer interactions according to the parameter $D_x$, represented as a probability, which governs the strength of coupling across layers.
The parameter $D_x$ acts as a superdiffusion parameter: higher values enhance cross-layer coupling and accelerate diffusion.
As shown in \cite{gomez2013diffusion}, by analysing the second smallest eigenvalue $\lambda_2$ of the supra-Laplacian matrix, one can assess whether the coupling strength $D_x$ facilitates faster diffusion across the multiplex network compared to individual layers.
This enables analytical tuning of $D_x$ to describe superdiffusive dynamics.
\\ \ \\
\noindent \textbf{Network Representations}. The data used to represent the mental lexicon as a cognitive multiplex network are shared between Case Study 2 and Case Study 3. Instead, in Case Study 1, we use a single-layer, semantic network, following the approach described in \cite{ciringione2024math}.

\textit{Network Data for Case Study 1}. We use behavioural forma mentis networks \cite{stella2019forma} to reconstruct the network of memory recalls produced by students with high and low math anxiety, as captured by a psychometric approach (Math Anxiety Scale) \cite{ciringione2024math}. Behavioural forma mentis networks are obtained from valence norms and free association data collected from 72 Italian undergraduate psychology students and from 300 GPT-simulated students. For these simulations, OpenAI's GPT 3.5 was used, as in the past work by Ciringione and colleagues \cite{ciringione2024math}. To build the free associations in forma mentis networks, the authors \cite{ciringione2024math} designed a task consisting of two parts: a free association game, and a valence evaluation. The task is roughly structured as the following instruction: \textit{Write a list of 3 words that come to your mind when you think of $CUE\_WORD$ and assign a valence score to each word as either  "positive", "negative", or "neutral"}.
Cue words are STEM-related, e.g., they include concepts such as \textit{mathematics}. 
A network $\mathcal{G}$ is then constructed using the word lists provided by each participant, connecting with an edge the cue word to each of the provided responses.
A statistical analysis of word valence is performed to assign a unique label to each word - either "positive", "negative", or "neutral" \cite{stella2019forma,ciringione2024math}. These labels are represented as colours in the mindset stream plots, cf. \cite{brian2023introducing} and Figure \ref{fig:mindset_streams}. 

\textit{Network Data for Case Studies 2 and 3}. We represent the mental lexicon as a multiplex network $\mathcal{M}$ \cite{stella2024cognitive}, where concepts are nodes connected via two layers, one for representing semantic associations and one for phonological similarities. The semantic layer is proxied by free associations - i.e., memory recall patterns of pairs of words - based on over 1 million responses from the Small World of Words (SWoW) dataset \cite{de2019small}. The phonological layer connects words based on phonological similarities, which are encoded here, as in past works \cite{vitevitch2008can,stella2017multiplex,stella2018multiplex}, by connecting any two words whose IPA (International Phonetic Alphabet) transcriptions differ by the addition, substitution or deletion of one phoneme only. By considering them in word forms (rather than IPA transcriptions), one can have the same set of nodes replicated across layers (see Figure \ref{fig:multiplex}). This alignment makes it possible to consider a multiplex network representation of the mental lexicon that was already investigated in past approaches \cite{stella2020multiplex}. Other approaches might rather represent word forms on a semantic layer and phonological IPA transcriptions on another layer, however this would lead to multilayer networks different from the one considered here. For the difference between multilayer and multiplex networks we refer the interested reader to \cite{stella2024cognitive}. To constrain simulations along nodes being connected on both the phonological and the semantic layers, we considered only the Largest Viable Cluster (LVC) of this multiplex representation, including $|V|=4118$ nodes and $|E|=26229$ edges. The LVC is the subgraph composed of all nodes for which exists a path in the multiplex network that does not exist in the layers separately, following \cite{stella2018multiplex, stella2020multiplex}. On this 2-layer representation we explore the inter-layer diffusion parameter $D_x$, which regulates superdiffusion, and it is shown in Figure \ref{fig:multiplex}D.


\section{Results}
\label{sec:results}

We test SpreadPy in 3 case studies, modelling (1) math anxiety patterns in human- and LLM-based behavioural data; (2) item-level difficulty in a creativity task; and (3) picture naming data from individuals with aphasia in psycholinguistic experiments. We treat each case study as independent from the others, with its own aims and scope, results and discussions.

\subsection{Case Study 1: SpreadPy and psychometric data from math anxiety}

\textbf{Main goal.} 
This section investigates activation patterns related to the concepts of "mathematics" and "statistics", and their connections with psychological variables such as anxiety, stress, and depression. Our model was used to simulate lexical activation across single-layer networks derived from different participant groups. This allowed us to identify and study group-based differences in the organization of concepts in these networks. Specifically, we compared students with high math anxiety (HMA) and low math anxiety (LMA) to examine how varying levels of math anxiety influence both the structure and the activation dynamics of conceptual networks. Particular attention was paid to how mathematical and statistical concepts connect with affective states such as anxiety, stress, and depression.

\textbf{Cognitive networks and math anxiety}.
Math anxiety is a multifaceted phenomenon that impairs cognitive functioning and emotional well-being, often manifesting as worry, tension, and avoidance in response to math-related situations \cite{hembree1990nature,ashcraft2002math}. Cognitive network science offers powerful tools to investigate how such anxiety is structured and sustained within the mental lexicon \cite{stella2022network}. Networks built from free association and valence attribution tasks can model this associative knowledge and capture how individuals mentally represent concepts related to science or mathematics and their affective tone \cite{stella2019forma,stella2022network}. 
In individuals with high math anxiety, such networks tend to show more fragmented associative structures and stronger links between math-related concepts negative emotions such as stress or fear \cite{stella2022network}. These associative patterns not only reflect memory recall but may also reinforce cognitive biases like avoidance behaviour, which contributes to the persistence of anxiety \cite{gangemi2012behavior,hayes1996experiential}. 
By modelling these associations as networks, researchers can analyse how affective content shapes knowledge structure and how anxiety constrains the cognitive complexity around the field of mathematics \cite{stella2022network}.

\begin{figure}[t!]
\centering
\includegraphics[scale=0.43]{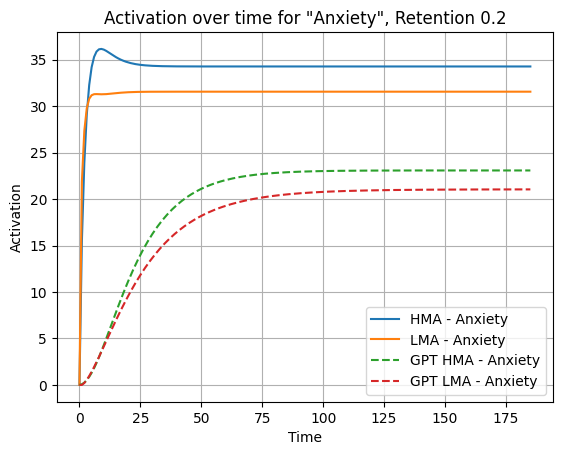}
\includegraphics[scale=0.43]{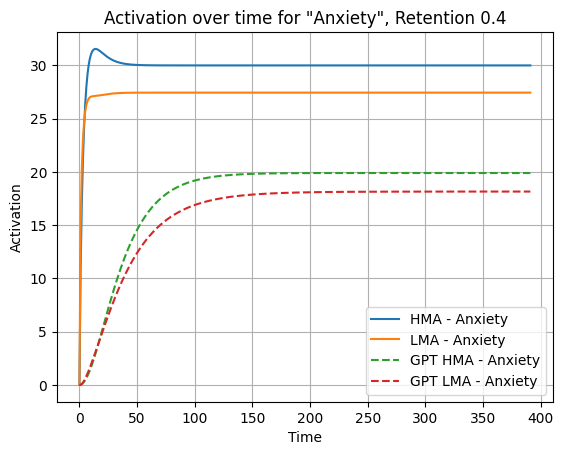}
\includegraphics[scale=0.43]{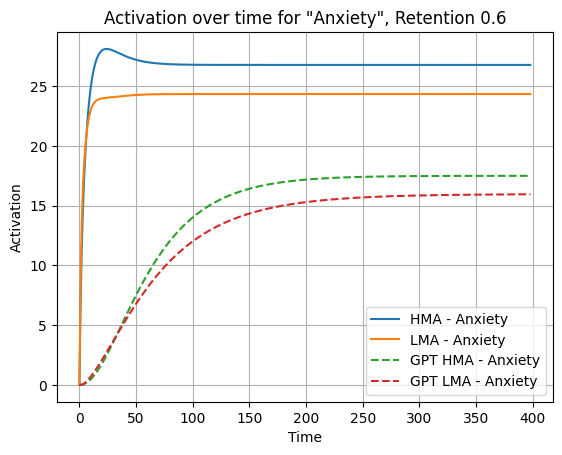}
\includegraphics[scale=0.43]{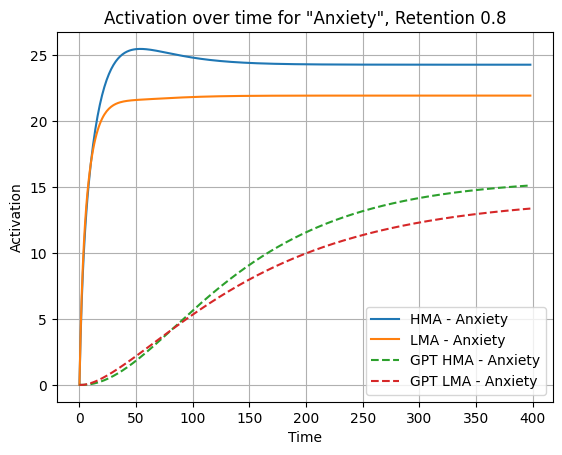}
\includegraphics[scale=0.43]{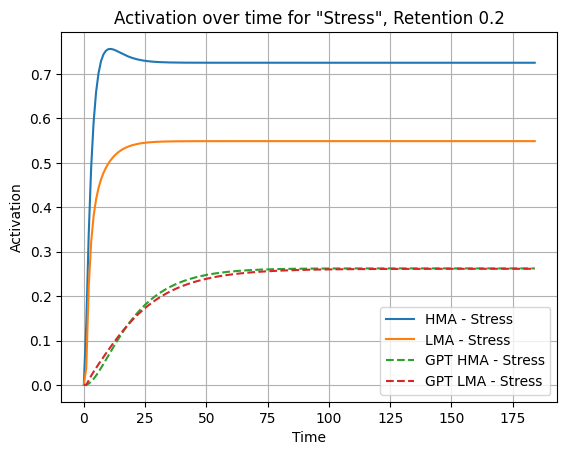}
\includegraphics[scale=0.43]{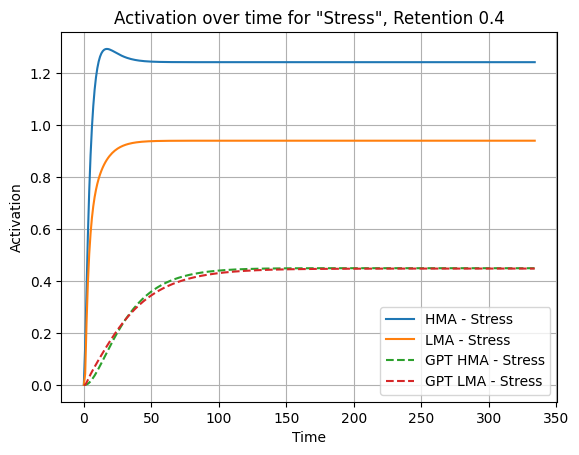}
\includegraphics[scale=0.43]{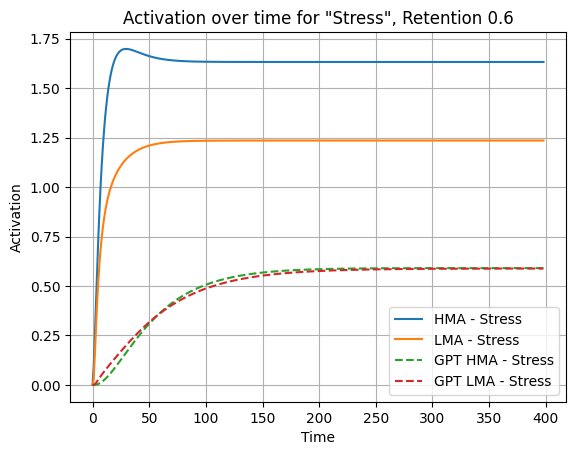}
\includegraphics[scale=0.43]{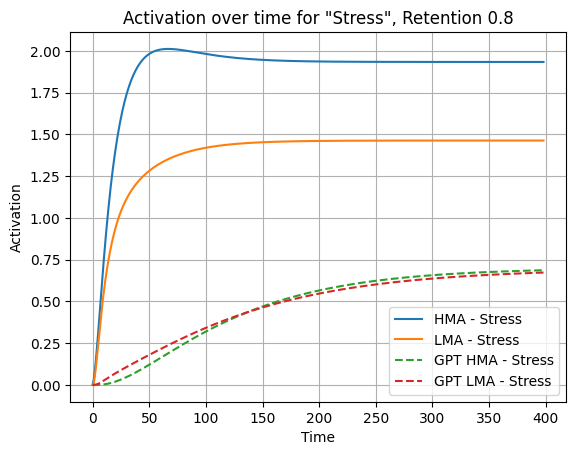}
\caption{Overall activation in HMA and LMA students and GPT-HMA and GPT-LMA students for the words "anxiety" (top 4 panels) and "stress" (bottom 4 panels) after the initial activation of the cue "mathematics" at different retention rates.}
\label{fig:activ_gpt}
\end{figure}

\begin{figure}[t!]
\centering
\subfloat[Path between "maths" and "anxiety" in HMA.]{\includegraphics[scale=0.25]{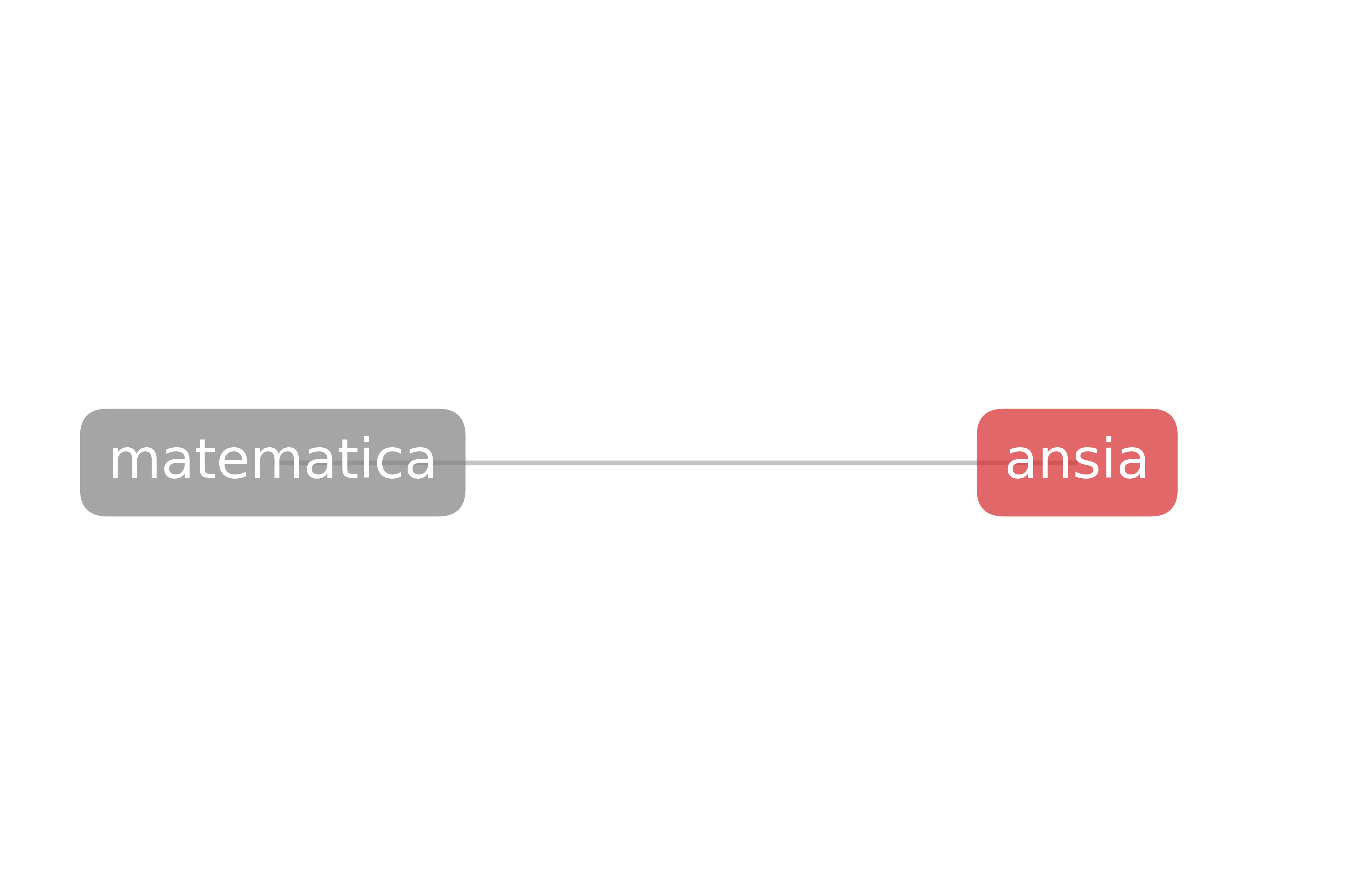}}
\subfloat[Path between "maths" and "anxiety" in LMA.]{\includegraphics[scale=0.25]{matematica_ansia_HMA.png}}
\qquad
\subfloat[Path between "maths" and "stress" in HMA.]{\includegraphics[scale=0.25]{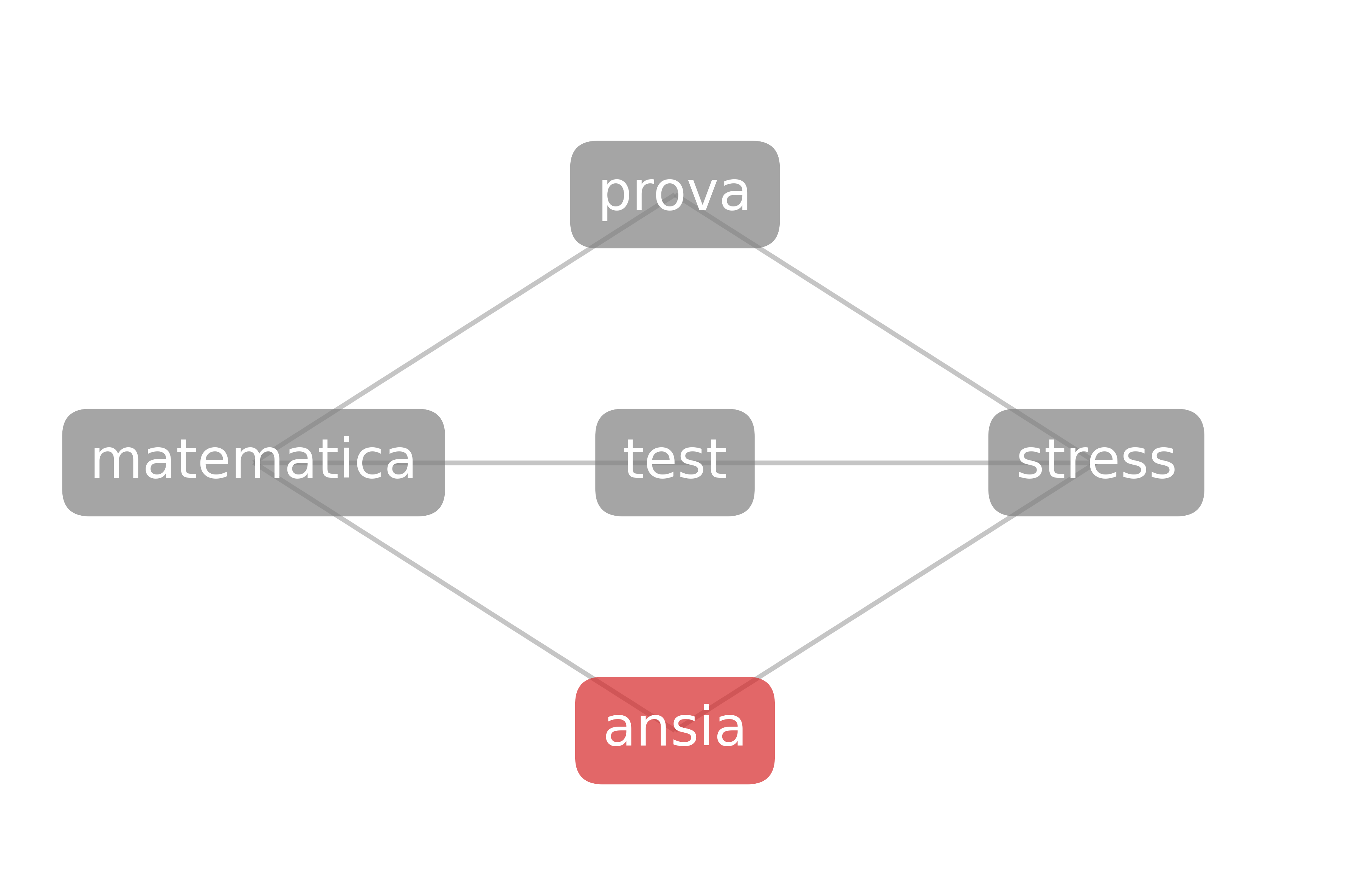}}
\subfloat[Path between "maths" and "stress" in LMA.]{\includegraphics[scale=0.25]{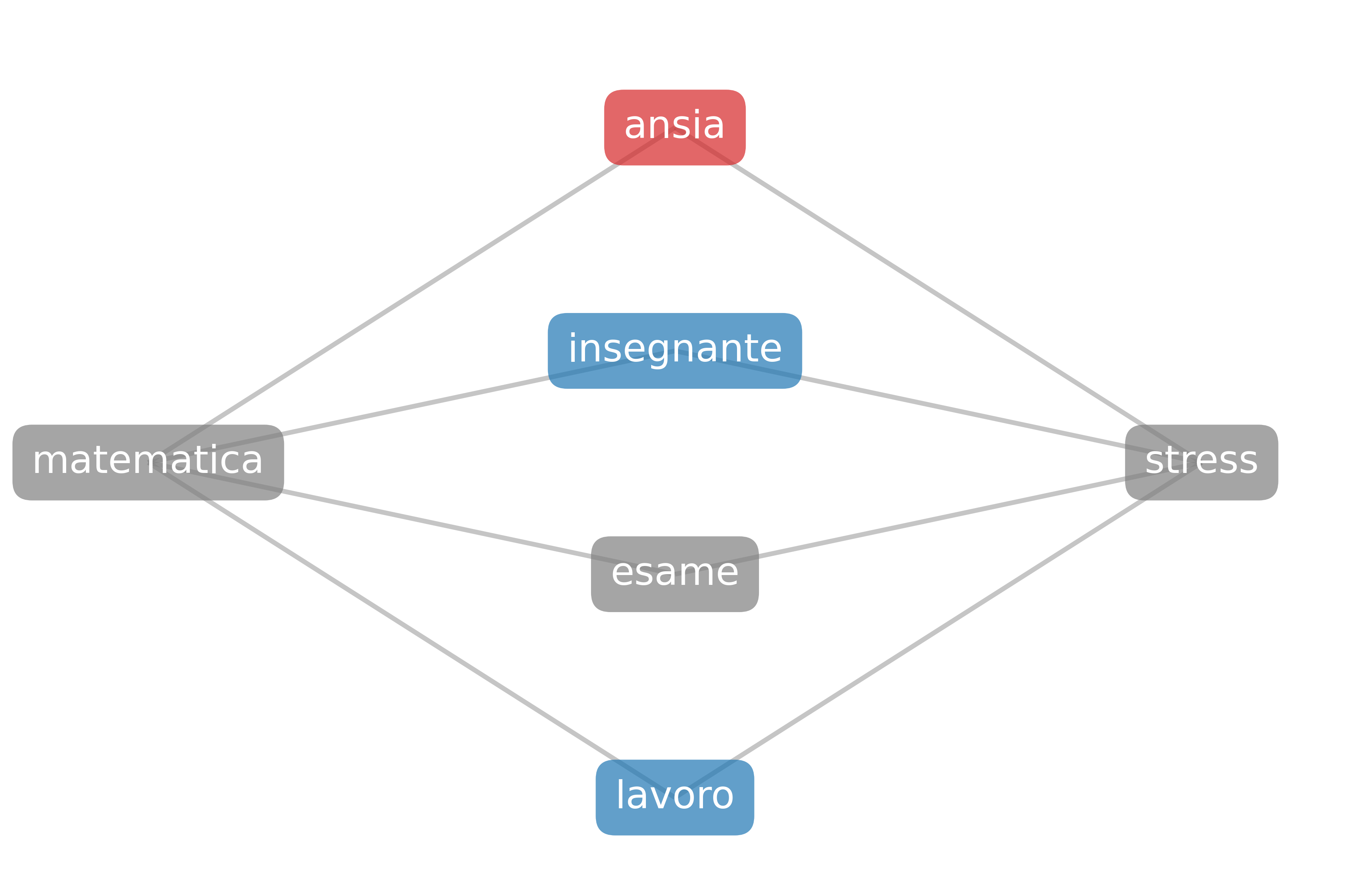}}
\qquad
\subfloat[Path between "maths" and "depression" in HMA.]{\includegraphics[scale=0.25]{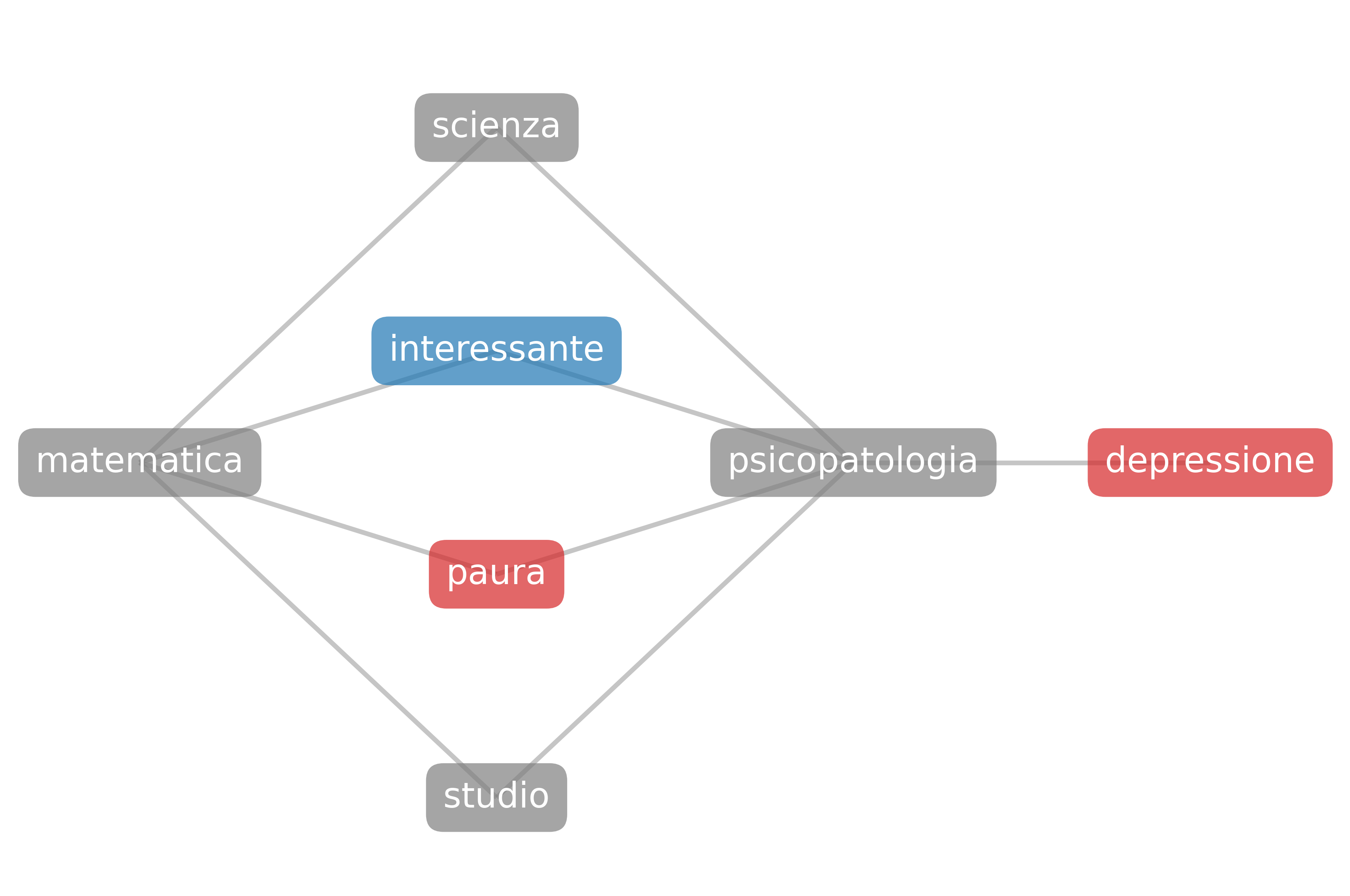}}
\subfloat[Path between "maths" and "depression" in LMA.]{\includegraphics[scale=0.25]{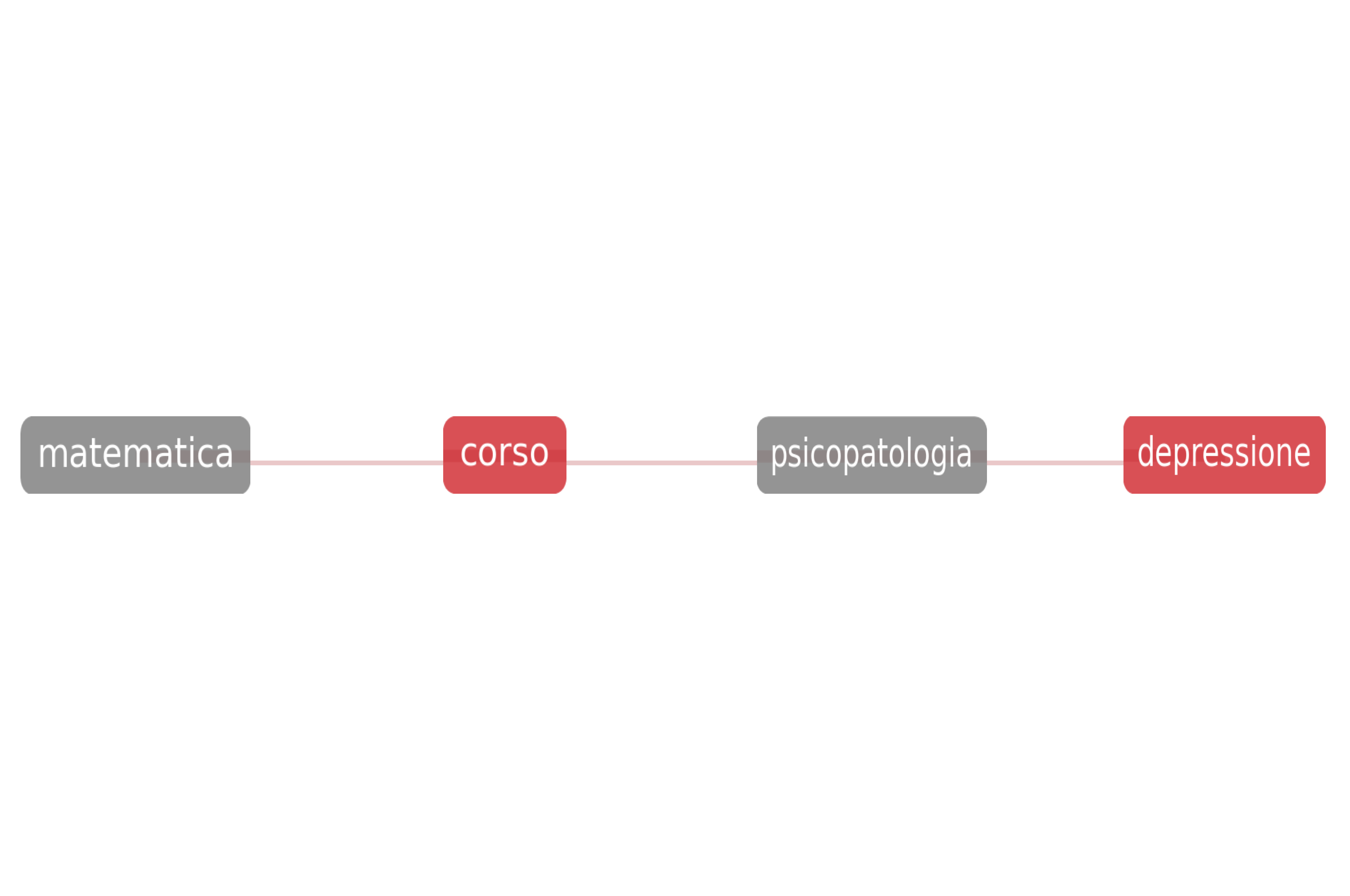}}
\caption{Mindset streams between "mathematics" (matematica) and "anxiety" (ansia), "stress", and "depression" (depressione) in HMA and LMA networks. The concepts are represented by Italian words because the networks were built from Italian data (see main text).}
\label{fig:mindset_streams}
\end{figure}

\textbf{Data and modelling setup}.
We simulated how the activation of the trigger node "mathematics" propagates through cognitive networks of interconnected concepts, affecting nodes associated with emotional states, such as stress, anxiety, and depression. 
To examine whether math anxiety alters associative pathways in cognitive networks, we simulated spreading activation from the node "mathematics" across group-based single-layer semantic networks. These networks were built from Italian free association data (see Section 2) collected from 72 Italian undergraduate psychology students, who were classified as having high (HMA) or low math axiety (LMA) based on the Mathematics Anxiety Scale. Participants responded to 40 cue words related to STEM, mental health, academic evaluation, and personal attitudes. Additional networks were generated by GPT-3.5 simulating students with high (GPT-HMA) or low math anxiety (GPT-LMA). 
For each of these four datasets (human HMA, human LMA, GPT-HMA, GPT-LMA) a separate single-layer network was built. 

SpreadPy was used to simulate spreading activation in these single-layer networks across different retention rates ($R = 0.2, 0.4, 0.6, 0.8$) to analyse possible differences in the propagation patterns; decay and suppress parameters were set to 0.
We then observed the spreading of activation through each network independently from one another and up to 400 time steps. We tracked the maximum activation level $\alpha_m$ and the time step of maximum activation $t_m$, representing the peak in activation following each triggered node. Furthermore, metrics such as node degree and shortest path lengths were collected as key elements of the structural configuration of the network.
The networks were then visualized through mindset streams, as described by Brian and Stella \cite{brian2023introducing}, which allow to highlight connections between two words in conceptual maps. These subgraphs are created from all the shortest paths between the concepts and allow associating linked nodes in a network to study their affective patterns. The valence of the connections within the stream is determined by the colour of the links:  blue for positive valence, red for negative valence, and gray for neutral concepts \cite{brian2023introducing}.

\textbf{Stronger and earlier activation of "anxiety" in high-math-anxiety students.} 
Figure \ref{fig:activ_gpt} illustrates the different activation levels for the nodes "anxiety" and "stress" across networks at various retention rates (0.2, 0.4, 0.6 and 0.8). When activation is injected into the node "mathematics" and allowed to diffuse, the HMA network drives the target word "anxiety" to its peak within the first ten time steps. The amplitude is $\approx 15-25\%$ higher than in the low-math-anxiety (LMA) network across all four retention regimes $(R = 0.2 → 0.8)$. The LMA curve rises more slowly and plateaus at a markedly lower level, indicating a weaker semantic-affective coupling between mathematics and anxiety in non-anxious students. The network itself displays an anxiety bias in HMA learners through the prevalence of negative associations given to "mathematics". Lower retention ($R = 0.2$) produces the steepest overshoot - activation in the high-math anxiety network peaks above 35 units (see Figure \ref{fig:activ_gpt}). As retention increases to $R = 0.8$, the curves flatten and converge on a common asymptote, yet the vertical gap between HMA and LMA remains around 5-6 units. This robustness implies that the difference is not a fragile artifact but a structural property of the students' associative knowledge structure. Replacing the probe with "stress" (Figure \ref{fig:activ_gpt} bottom four panels) yields peak amplitudes two orders of magnitude smaller and a much slower approach to equilibrium. Crucially, the HMA-LMA separation persists but contracts to a few tenths of an energy unit. The lexicon therefore encodes "mathematics" as explicitly \textit{anxiogenic} rather than generically negative: only the "anxiety" concept is strongly and quickly co-activated, while "stress" shows weak, delayed coupling.

\textbf{GPT-simulated students replicate the human activation hierarchy but with dampened intensity.} 
The GPT-simulated forma mentis networks successfully reproduce the overall rank ordering observed in human data: when activation spreads from the cue word "mathematics", the target concepts "anxiety" and "stress", respectively, reach their highest activation levels in the high math anxiety (HMA) group, followed by the low math anxiety (LMA) group ($HMA > LMA$). This same pattern holds for GPT-simulated participants: GPT-HMA consistently shows higher activation than GPT-LMA. However, the overall intensity of activation is markedly reduced in GPT-generated networks compared to human ones, and the diffusion unfolds more slowly over time. 
As shown in Figure \ref{fig:activ_gpt}, the GPT curves sit below the human counterparts and exhibit delayed and flatter growth, taking longer to reach peak activation. In some conditions, the difference between GPT-HMA and GPT-LMA becomes almost negligible (see Figure\ref{fig:activ_gpt}, bottom panels). This reduced contrast indicates that the models capture the directionality of affective links but underestimate their strength, suggesting that purely text-based learning misses experiential reinforcement present in real students.

\textbf{Mindset streams reveal different pathways from "mathematics" to "anxiety/stress/depression" in HMA and LMA networks.} 
Mindset streams are a collection of all shortest paths that join two concepts inside a forma mentis network \cite{brian2023introducing}. They provide a graphical thought-traces, exposing both the semantic bridges and the emotional tone (valence) that link those ideas. Figure \ref{fig:mindset_streams} (a) shows a short mindset stream linking "mathematics" and "anxiety" through a single direct connection, indicating a strong and immediate association between the two concepts. In contrast, Figure \ref{fig:mindset_streams} (e) illustrates a longer stream - three links connecting "mathematics" to "depression" - suggesting a more distant relationship that relies on intermediate concepts. In this way, mindset streams help reveal how directly concepts are associated and which intermediary ideas bridge them \cite{brian2023introducing}. 

Applied to the human networks of highly anxious and lowly anxious students, the streams show that "mathematics $\rightarrow$ anxiety" is a one-step direct connection in both groups. This indicates that the subject of mathematics alone already evokes apprehension in students. In contrast, the paths from "mathematics" to "stress" diverge: In HMA students the path runs through "test/attempt $\rightarrow$ anxiety", pinpointing test anxiety as the main channel. In LMA students it detours via "job $\rightarrow$ exam $\rightarrow$ teacher", mixing neutral and positive cues and signalling a milder valence. Paths to "depression" are longest: in HMA they wind through academically flavoured but largely neutral terms ("science $\rightarrow$ study/interesting/fear/study $\rightarrow$ psychopathology"), and in LMA they are even more indirect, making depression cognitively distant from mathematics. Together, these condensed insights indicate a tight, direct coupling between mathematics and anxiety, whereas links to "stress" and "depression" are weaker and context-dependent, with test-related notions amplifying distress in the high-anxiety group.

\textbf{Discussion of results.} 
The simulations reveal a strong and immediate semantic link between "mathematics" and "anxiety" in the networks of high-math-anxiety (HMA) students. This suggests that interventions should go beyond general reduction of stress-related vocabulary and instead focus on weakening this direct associative pathway. For example, this could be achieved by embedding more positive or neutral concepts between "mathematics" and "anxiety" in the students' mental lexicon. 
The spreading activation model not only captures these affective trajectories in detail, but also offers a measurable framework to evaluate whether psychological or pedagogical interventions, such as mindset training or curricular changes, have effectively altered the structure and emotional tone of students' conceptual networks related to "mathematics" and "anxiety".

\begin{figure}[t!]
\centering
\includegraphics[scale=0.62]{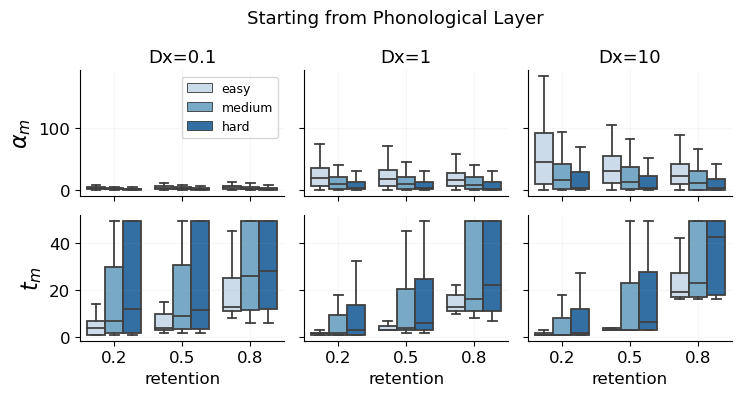}
\includegraphics[scale=0.62]{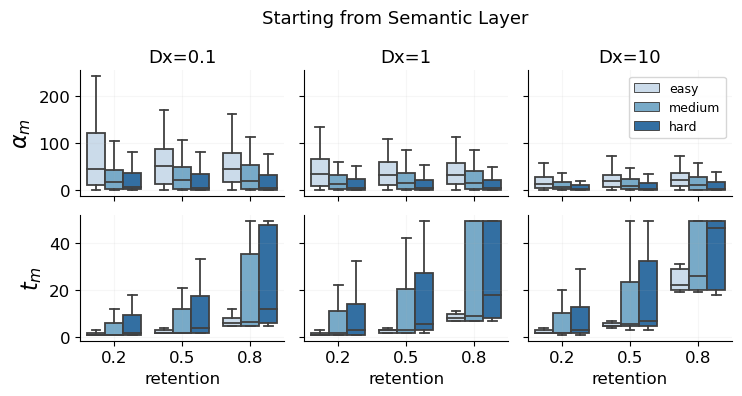}
\includegraphics[scale=0.62]{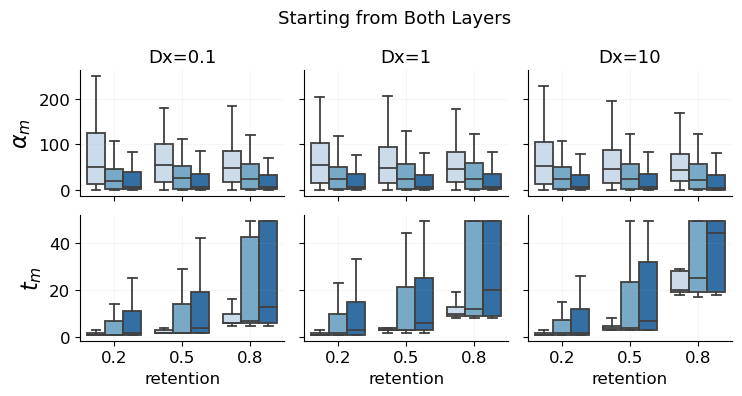}
\caption{\textit{Creativity}: Boxplots displaying maximum activation ($\alpha_m$) and time to reach it ($t_m$) at different diffusion ($Dx=0.1, 1, 10)$ and retention ($R=0.2, 0.5, 0.8$) rates, starting from different layers when diffusion is seeded (phonological or semantic replica, or both).}
\label{fig:creativity}
\end{figure}

\subsection{Case Study 2: Modelling creativity data with SpreadPy}

\textbf{Main goal.} 
This case study assesses whether the SpreadPy library can predict item-level difficulty in a classic creativity task, the Remote Associates Test (RAT) \cite{mednick1962associative}, by simulating activation dynamics on a multiplex network. We investigated whether different patterns of activation spreading across semantic and phonological layers could explain why some RAT items are easier to solve than others. Specifically, we asked whether the two activation-based outcomes maximum activation ($\alpha_m$) and time of maximum activation ($t_m$) could help differentiate between easy, medium and hard difficult RAT items. We further tested how this relationship is modulated by key parameters of the SpreadPy model, including retention rate, diffusion regime (standard vs. superdiffusion) and initial activation layer (semantic, phonological or both).

\textbf{Creativity and multiplex networks}. 
Creativity research has increasingly adopted tools from network science to model how people generate novel ideas by connecting remote concepts \cite{kenett2019semantic,beaty2023semantic}. The Remote Associates Test (RAT) was developed by Mednick \cite{mednick1962associative} and is widely used to assess convergent creativity. Solving the RAT evaluates the ability to identify a word that links three seemingly unrelated cues (e.g. cottage - Swiss - cake --> cheese).
During the test, participants must often inhibit misleading initial associations and rely on more distant, non-obvious connections. This makes the RAT particularly sensitive to the structure of the mental lexicon \cite{bowers1990intuition,bowden2003normative}. 
Especially difficult RAT items typically require longer and more indirect associative chains than easy items, retrieving a common target through less obvious associative pathways. This has made the RAT a valuable tool for studying insight problem solving, associative thinking, and creative cognition \cite{bowden2003normative,marko2019remote,wu2020systematic}.

While spreading activation has been proposed as a relevant cognitive mechanism in creative performance, few models have implemented this framework over multiplex networks. Prior studies have used single-layer networks to simulate activation spreading through the semantic network \cite{siew2019spreadr}. In contrast, SpreadPy allows us to simulate time-sensitive activation diffusion over a multiplex network. To our knowledge, this is the first simulation study to investigate how activation dynamics on a semantic-phonological multiplex network relate to difficulty levels of RAT items. 

\textbf{Data and modelling setup}. We selected a subset of established RAT items for which difficulty ratings were available \cite{bowers1990intuition,bowden2003normative,lee2014measure,wu2020systematic}. To ensure compatibility with our multiplex network, we retained only items for which all four relevant words (three cues and one target solution) were present in both network layers. This constraint was necessary because the absence of any node would disrupt the spreading process across the multiplex network. In each simulation, activation was seeded at the three cue nodes and allowed to spread across both semantic and phonological layers. We monitored only the solution node, recording two outcome variables per item: maximum activation ($\alpha_m$) and time of maximum activation ($t_m$). 

\textbf{Research hypothesis}. Our central hypothesis is that easier RAT items should allow the solution node to accumulate activation more quickly (lower $t_m$) and result in higher maximum activation (higher $\alpha_m$). Conversely, difficult items were expected to show slower (higher $t_m$) and weaker (smaller $\alpha_m$) activation build-up. This is hypothesized to occur due to greater conceptual distance between cues and target, as well as more interfering and misleading concepts in between. To further understand how item difficulty influences the activation patterns, we varied key simulation parameters to analyse their effect on $\alpha_m$ and $t_m$. We examined the influence of these three parameters on the activation patterns (see Figure \ref{fig:creativity}): (i) Retention rate (set to 0.2, 0.5, or 0.8); (ii) Diffusion mode (standard diffusion vs. superdiffusion); (iii) Initial activation layer (phonological only, semantic only, or both layers simultaneously).

\renewcommand\baselinestretch{1}\selectfont
\begin{table}[t!]
    \centering
    \caption{\textit{Creativity}: Cohen's d values for different starting layers. Bold values indicate statistically significant Kruskal-Wallis results (p<0.05).}
    \label{tab:cohen_d_creativity}
 
    \renewcommand{\arraystretch}{0.92} 
    \setlength{\tabcolsep}{7pt} 
    
    \begin{tabular}{l|l|ccc|ccc}
        \toprule
        \textbf{Dx} & \textbf{Groups} & \multicolumn{3}{c|}{\textbf{$\alpha_m$}} & \multicolumn{3}{c}{\textbf{$t_m$}} \\
        \midrule
        & & \textbf{R=0.2} & \textbf{R=0.5} & \textbf{R=0.8} & \textbf{R=0.2} & \textbf{R=0.5} & \textbf{R=0.8} \\
        \rowcolor{orange!10} 0.1 & easy vs hard & \textbf{0.27} & \textbf{0.38} & \textbf{0.43} & \textbf{0.69} & \textbf{0.72} & \textbf{0.57} \\
        \rowcolor{orange!10} 0.1 & easy vs medium & \textbf{0.19} & \textbf{0.29} & \textbf{0.33} & \textbf{0.46} & \textbf{0.55} & \textbf{0.52} \\
        \rowcolor{orange!10} 0.1 & medium vs hard & 0.06 & \textbf{0.10} & \textbf{0.14} & 0.26 & 0.23 & 0.08 \\
        \rowcolor{purple!10} 1 & easy vs hard & \textbf{0.78} & \textbf{0.76} & \textbf{0.72} & \textbf{0.53} & \textbf{0.55} & \textbf{0.62} \\
        \rowcolor{purple!10} 1 & easy vs medium & \textbf{0.48} & \textbf{0.46} & \textbf{0.44} & 0.40 & 0.42 & 0.47 \\
        \rowcolor{purple!10} 1 & medium vs hard & 0.27 & 0.29 & 0.29 & 0.18 & 0.19 & 0.16 \\
        \rowcolor{blue!10} 10 & easy vs hard & \textbf{0.82} & \textbf{0.79} & \textbf{0.74} & \textbf{0.37} & \textbf{0.48} & \textbf{0.61} \\
        \rowcolor{blue!10} 10 & easy vs medium & \textbf{0.51} & \textbf{0.45} & \textbf{0.42} & 0.22 & 0.35 & 0.38 \\
        \rowcolor{blue!10} 10 & medium vs hard & 0.26 & 0.30 & 0.30 & 0.17 & 0.15 & 0.23 \\
        \bottomrule
    \end{tabular}
       \subcaption{Starting From Phonological Layer}

    \begin{tabular}{l|l|ccc|ccc}
        \toprule
        \textbf{Dx} & \textbf{Groups} & \multicolumn{3}{c|}{\textbf{$\alpha_m$}} & \multicolumn{3}{c}{\textbf{$t_m$}} \\
        \midrule
        & & \textbf{R=0.2} & \textbf{R=0.5} & \textbf{R=0.8} & \textbf{R=0.2} & \textbf{R=0.5} & \textbf{R=0.8} \\
        \rowcolor{orange!10} 0.1 & easy vs hard & \textbf{0.81} & \textbf{0.82} & \textbf{0.78} & \textbf{0.43} & \textbf{0.46} & \textbf{0.64} \\
        \rowcolor{orange!10} 0.1 & easy vs medium & \textbf{0.50} & \textbf{0.49} & \textbf{0.45} & 0.30 & 0.25 & 0.45 \\
        \rowcolor{orange!10} 0.1 & medium vs hard & 0.23 & 0.28 & 0.30 & 0.17 & 0.26 & 0.22 \\
        \rowcolor{purple!10} 1 & easy vs hard & \textbf{0.82} & \textbf{0.80} & \textbf{0.77} & \textbf{0.49} & \textbf{0.49} & \textbf{0.65} \\
        \rowcolor{purple!10} 1 & easy vs medium & \textbf{0.50} & \textbf{0.47} & \textbf{0.43} & 0.33 & 0.33 & 0.44 \\
        \rowcolor{purple!10} 1 & medium vs hard & 0.26 & 0.29 & 0.31 & 0.21 & 0.20 & 0.21 \\
        \rowcolor{blue!10} 10 & easy vs hard & \textbf{0.69} & \textbf{0.74} & \textbf{0.73} & \textbf{0.44} & \textbf{0.57} & \textbf{0.59} \\
        \rowcolor{blue!10} 10 & easy vs medium & \textbf{0.37} & \textbf{0.43} & \textbf{0.42} & 0.34 & 0.45 & 0.36 \\
        \rowcolor{blue!10} 10 & medium vs hard & 0.28 & 0.29 & 0.30 & 0.13 & 0.16 & 0.23 \\
        \bottomrule
    \end{tabular}
     \subcaption{Starting From Semantic Layer}

    \begin{tabular}{l|l|ccc|ccc}
        \toprule
        \textbf{Dx} & \textbf{Groups} & \multicolumn{3}{c|}{\textbf{$\alpha_m$}} & \multicolumn{3}{c}{\textbf{$t_m$}} \\
        \midrule
        & & \textbf{R=0.2} & \textbf{R=0.5} & \textbf{R=0.8} & \textbf{R=0.2} & \textbf{R=0.5} & \textbf{R=0.8} \\
        \rowcolor{orange!10} 0.1 & easy vs hard & \textbf{0.81} & \textbf{0.82} & \textbf{0.78} & \textbf{0.51} & \textbf{0.52} & \textbf{0.64} \\
        \rowcolor{orange!10} 0.1 & easy vs medium & \textbf{0.51} & \textbf{0.50} & \textbf{0.46} & 0.40 & 0.42 & 0.50 \\
        \rowcolor{orange!10} 0.1 & medium vs hard & 0.23 & 0.28 & 0.29 & 0.16 & 0.14 & 0.17 \\
        \rowcolor{purple!10} 1 & easy vs hard & \textbf{0.81} & \textbf{0.80} & \textbf{0.76} & \textbf{0.52} & \textbf{0.53} & \textbf{0.64} \\
        \rowcolor{purple!10} 1 & easy vs medium & \textbf{0.50} & \textbf{0.47} & \textbf{0.44} & 0.38 & \textbf{0.41} & 0.46 \\
        \rowcolor{purple!10} 1 & medium vs hard & 0.27 & 0.29 & 0.30 & 0.18 & 0.17 & 0.18 \\
        \rowcolor{blue!10} 10 & easy vs hard & \textbf{0.80} & \textbf{0.77} & \textbf{0.74} & \textbf{0.44} & \textbf{0.54} & \textbf{0.60} \\
        \rowcolor{blue!10} 10 & easy vs medium & \textbf{0.49} & \textbf{0.45} & \textbf{0.42} & \textbf{0.29} & 0.41 & 0.37 \\
        \rowcolor{blue!10} 10 & medium vs hard & 0.27 & 0.30 & 0.30 & 0.19 & 0.17 & 0.23 \\
        \bottomrule
    \end{tabular}
    \subcaption{Starting From Both Layers}
\end{table}
\renewcommand\baselinestretch{1.5}\selectfont

\renewcommand\baselinestretch{1}\selectfont
\begin{table}[t!]
\centering
\caption{Parameter estimates for the multinomial logistic regression using the \textit{easy} class as baseline}
\begin{tabular}{lllrrrrl}
\hline
Variable & Category & Estimate & Std. Error & z-value & p-value & Significance \\
\hline
Intercept & hard & 0.6523 & 0.0566 & 11.516 & $<$0.0001 & *** \\
Intercept & medium & 0.0938 & 0.0647 & 1.450 & 0.1470 &  \\
r & hard & -0.7475 & 0.1588 & -4.706 & $<$0.0001 & *** \\
r & medium & -0.5731 & 0.1793 & -3.196 & 0.0014 & ** \\
$\alpha_{max}$ & hard & -0.0139 & 0.0011 & -12.813 & $<$0.0001 & *** \\
$\alpha_{max}$ & medium & -0.0069 & 0.0011 & -6.362 & $<$0.0001 & *** \\
$t_{max}$ & hard & 0.0255 & 0.0028 & 9.079 & $<$0.0001 & *** \\
$t_{max}$ & medium & 0.0201 & 0.0031 & 6.402 & $<$0.0001 & *** \\
$s=phon$ & hard & 0.2582 & 0.0505 & 5.114 & $<$0.0001 & *** \\
$s=phon$ & medium & -0.0172 & 0.0578 & -0.297 & 0.7662 &  \\
$s=sem$ & hard & 0.4329 & 0.0538 & 8.042 & $<$0.0001 & *** \\
$s=sem$ & medium & 0.1201 & 0.0615 & 1.953 & 0.0508 & . \\
$S$ & hard & 0.0706 & 0.0080 & 8.769 & $<$0.0001 & *** \\
$S$ & medium & 0.0279 & 0.0091 & 3.060 & 0.0022 & ** \\
\hline
\multicolumn{7}{l}{Significance codes: *** $p<0.001$, ** $p<0.01$, * $p<0.05$, . $p<0.1$}
\end{tabular}
\label{tab:multinom}
\end{table}
\renewcommand\baselinestretch{1.5}\selectfont

\textbf{Single-layer versus multiplex diffusion.} SpreadPy enables diffusion simulations either on individual layers, see Table \ref{tab:cohen_d_creativity}, or on the whole multiplex structure, see Table \ref{tab:creativity_single_layer}. When separating difficulty levels in terms of $\alpha_m$ or $t_m$, the strongest effect sizes are found in terms of multiplex diffusion ($d\approx0.8$ for easy vs. hard, $p<0.05$), either when starting from the semantic layer or when activating both replicas or words on both layers. When diffusion is restricted only on the semantic layer, differences between RAT difficulty levels decrease while remaining statistically significant ($d\approx0.6$ for easy vs. hard, $p<0.05$). Interestingly, diffusion on the phonological layer only (see Table \ref{tab:creativity_single_layer}) never gives rise to statistically significant differences. These numerical evidences indicate that: (i) the RAT difficulty levels, reflecting varying creativity, encode features reflected by both semantic and phonological similarities between words; (ii) semantic features are predominant, compared to phonological features, when determining the difficulty of searching for solutions in a RAT task. The first point motivates us further to explore multiplex diffusion with the RAT creativity data.

\textbf{Peak magnitude outperforms activation timing in separating difficulty levels.} Across all simulation conditions, maximum activation ($\alpha_m$) provided the clearest separation of RAT item difficulty. This was especially evident in comparisons of easy vs. hard items (averaged Cohen's d across diffusion regimes and retention rates of 0.73), but also visible in easy versus medium items (averaged cohen's $d = 0.44$). By contrast, the time to reach maximum activation ($t_m$) showed fewer significant contrasts ($p<0.05$), with only easy versus hard comparisons consistently reaching moderate effect sizes (averaged Cohen's d of 0.55). Comparisons of medium versus hard items were generally not significant ($p>0.05$) for both $\alpha_m$ and $t_m$.
These results suggest that peak activation strength, rather than retrieval speed, may better reflect the associative accessibility of a correct solution in a convergent creative task such as the RAT.

\textbf{Easier items reach higher peak activation.} As visible in Figure \ref{fig:creativity}, maximum activation levels were significantly higher for easy RAT items compared to medium and hard items, across all retention and diffusion conditions. When starting the activation on the phonological layer (Figure \ref{fig:creativity}, top panel) there are only small effects under multiplex diffusion ($D_x=0.1$; mean $d=0.27$ for easy vs medium; mean $d=0.36$ for easy vs hard). Under superdiffusion ($D_x=1, 10$) and when starting from either the semantic layer or both layers simultaneously, effect sizes increase to moderate or strong with mean $d=0.46$ (easy vs medium) and mean $d=0.77$ (easy vs strong). Noticeably, when cue words were activated on both layers simultaneously, differences between diffusion modes decreased and effect sizes became more stable (see Table \ref{tab:cohen_d_creativity}). 
No reliable differences were found between medium and hard items.

\textbf{Easier items reach peak activation faster.} Throughout the simulations, easy RAT items consistently reached their peak activation earlier (lower $t_m$) than medium or hard items. This pattern stayed consistent across different starting layers, retention rates and diffusion regimes. This indicates that retrieval latency, i.e. the time it takes to retrieve an item from semantic memory which is here captured by $t_m$ (time until maximum activation), is a valuable indicator of semantic accessibility \cite{collins1969retrieval}. 
The comparison of easy vs hard items consistently showed significant differences with moderate effect sizes (mean $d=0.55$). Differences between easy and medium items were inconsistent, with significant findings only in some conditions (phonological seed layer, $D_x=0.1$; both seed layers, $D_x=1$, $R=0.5$; both seed layers, $D_x=10$, $R=0.2$) with weak to moderate effect sizes (mean $d=0.44$). 
Taken together, these findings indicate that $t_m$ robustly distinguishes easy from hard RAT items, regardless of seeding condition, retention rate and diffusion strength. 

\textbf{Retention rates modulate activation timing, not activation magnitude.} When changing retention rates (0.2; 0.5; 0.8) in the simulations, maximum activation levels remained largely unaffected. This was somewhat unexpected, considering a higher retention rate of 0.8 means a node retains 80 percent of ingoing activation for itself, spreading only 20 percent to its neighbouring nodes. In that case, less activation should keep spreading throughout the network, resulting in potentially less activation accumulating at the solution node. Thus, one might expect to observe higher maximum activation levels for lower retention rates (more activation is spread on) than for higher retention rates (less activation is spread on). However, even though we found some indications for such a pattern, our results indicate no strong variations in maximum activation levels according to modulations in retention rate.

In contrast, we found considerable correlations between retention values and the time it took to reach peak activation. Across all diffusion regimes and starting layers, higher retention consistently increased the latency of maximum activation. This effect was especially pronounced in the superdiffusion regime ($D_x=10$), where increasing retention from 0.2 to 0.8 prolonged average and minimum $t_m$ values for all three item difficulty types. For example, when seeding the activation at the phonological layer under superdiffusion ($D_x=10$), the contrast in $t_m$ between easy and hard items increased from $d=0.37$ (at $R= 0.2$) to $d = 0.61$ at ($R= 0.8$), while corresponding $\alpha_m$ differences remained stable (mean $d = 0.78$). This decoupling suggests that retention does not alter whether a solution becomes accessible (i.e. if it crosses a certain activation threshold), but when it becomes accessible (i.e. how fast activation can accumulate at the solution node). In behavioural terms, this suggests a failure to access a solution node might arise not due to insufficient activation levels, but due to a temporal delay, especially when participants are under time pressure \cite{collins1969retrieval,maril2001tip}.

\textbf{Reversed patterns across phonological and semantic layers.} We find reversed patterns for maximum activation $\alpha_m$ when modulating the diffusion parameter  and when starting activation either from the phonological or the semantic layer. When seeding the activation at the phonological layer, nearly no difficulty-related differences emerge at multiplex diffusion ($D_x=0.1$) with small effect sizes (mean Cohen's $d=0.26$ for maximum activation). This indicates that spreading activation confined to phonological similarity alone fails to capture the semantic associations necessary for solving the Remote Associates Task. However, under superdiffusion ($D_x=10$), phonological cues gain access to the semantic layer via enhanced inter-layer coupling, allowing them to quickly \textit{hop} into the semantic layer, resulting in larger effect sizes (mean Cohen's $d=0.62$ for maximum activation). This suggests that phonological structure alone does not encode RAT difficulty, but rather becomes informative when it allows shortcuts into the semantic network.

Conversely, when seeding the activation at the semantic layer, we find the strongest separation of item difficulty in the result variable of maximum activation at multiplex diffusion ($D_x=0.1$) with moderate Cohen's d values (mean $d=0.64$ for maximum activation). Allowing activation to diffuse excessively into the phonological layer under superdiffusion ($D_x=10$) reduces these effect sizes (mean $d=0.56$). Keeping activation within the semantic layer seems to preserve the network's capacity to discriminate easy versus hard items, whereas adding excessive diffusion into the phonological layer blurs that discrimination. 
This pattern aligns with the reliance of the RAT on semantic, not phonological, associations \cite{mednick1962associative,bowden2003normative}. While phonological activation may facilitate access to the solution under certain conditions, it does not appear to encode RAT item difficulty as effectively or directly as the semantic layer.

\textbf{Multinomial logistic regression complements size analysis.} We ran a multinomial logistic regression to identify how different predictors influence the likelihood of the categories \textit{hard} and \textit{medium}, relative to a baseline category (\textit{easy}). Significant negative coefficients for $r$ (retention rate) and $\alpha_{max}$ (maximum activation level) indicate that higher values on these predictors decrease the odds of responses being classified as \textit{hard} or \textit{medium}, with particularly strong effects for \textit{hard}. Conversely, significant positive coefficients for $t_{max}$ (time of maximum activation), $s=sem$ (start from the semantic layer), $s=phon$ (start from the phonological layer), and $S$ (superdiffusion parameter) indicate that increases in these predictors enhance the likelihood of responses being categorized as \textit{hard} or, to a lesser extent, \textit{medium}. The lack of significant effect for $s=phon$ in predicting the \textit{medium} category indicates this predictor primarily differentiates between \textit{hard} and \textit{easy} categories. Overall, predictors show consistently stronger and highly significant effects in distinguishing \textit{hard} responses relative to the baseline category, thus agreeing with the results on size analysis.

\textbf{Discussion of results.} Our simulations demonstrate that easier RAT items consistently reached higher peak activation levels (higher $\alpha_m$) and did so more quickly (lower $t_m$) than medium and hard items, indicating that easy items were structurally more accessible. This is supported by the spreading activation theory, which posits that nodes closer to the initial activated node accumulate activation more rapidly than remote ones \cite{collins1975spreading}. The distinction between easy and harder items was robust across parameters, suggesting that temporal and magnitude-based metrics like $t_m$ and $\alpha_m$ are valid indicators of structural accessibility. However, the lack of distinction between medium and hard items implies that standard RAT difficulty levels may reflect a binary division of "readily accessible" vs. "remote" rather than a fine-grained continuum.
Retention analyses revealed that higher retention delays peak activation without affecting its strength. This finding highlights the importance of temporal constraints in retrieval of a solution node \cite{collins1969retrieval,maril2001tip}. In practical terms, participants may fail to retrieve a correct solution not because activation never reaches a threshold, but because it does so too late under time-limited conditions.
Phonological seeding only aided discrimination under superdiffusion, highlighting that RAT performance primarily relies on semantic pathways. This aligns with previous research, identifying semantic network structure as a core driver of creative problem solving during verbal creativity tasks like the RAT \cite{mednick1962associative,bowden2003normative,lee2014measure}.

Future studies should examine whether individual differences in retrieval speed and brain activation patterns can predict item difficulty and creativity measures, linking network dynamics to real-world creative performance \cite{beaty2023semantic,kenett2019semantic}.

\subsection{Case Study 3: Modelling aphasic data with SpreadPy}

\textbf{Main goal.} This case study assessed whether SpreadPy can reproduce and quantitatively dissociate the three canonical picture-naming errors observed in anomic aphasia: semantic, formal (phonological) and mixed paraphasias. Specifically, we asked whether diffusion dynamics on a multiplex lexical network predict empirical error frequencies and their characteristic types (formal, semantic or mixed), and which combinations of retention (R) and inter-layer coupling ($D_x$) maximize diagnostic separation.

\noindent \textbf{Multiplex networks and aphasia.} Aphasia syndromes can come with various deficits, such as anomia, which describes problems with accessing a desired word. In anomic aphasia, patients usually only suffer from word retrieval problems while other linguistic domains remain largely unimpaired. For example, a patient with anomic aphasia could struggle during a simple picture naming task, show slower reaction times and increased chances for mistakes, such as saying "waterfall" for a picture reporting a "bottle" \cite{kohn1985picture,harnish2018anomia}. 
Psycholinguists can search for patterns in mistaken productions within picture naming tasks, labelling mistakes as predominantly semantic, formal/phonological, or mixed, among other types of mistakes (e.g. no production at all, non-word) \cite{kohn1985picture,wilshire2002aphasic,kittredge2008effect}. Semantic mistakes take place when individuals can understand some semantic features of the target word - the word represented in the picture - but cannot wholly activate, retrieve or produce it entirely. Consequently, the produced mistake shares some semantic similarity with the desired target. An example of a semantic mistake would be naming a picture with a "table" as "supporter" \cite{kittredge2008effect}. Analogously to semantic mistakes being influenced by semantic similarities, formal or phonemic mistakes occur when individuals produce similar sounding words to the desired target. If picture naming is performed verbally, formal mistakes are mostly driven by phonological sound similarities. An example could be naming a picture showcasing a "bunny" as "funny" \cite{wilshire2002aphasic}. Mixed mistakes are relative to experimenters identifying both patterns of semantic and/or phonological similarities, e.g. saying "rat" for "cat". Identifying the types and rates of these mistakes can help psycholinguists to better understand the extent and severity of anomic aphasia within a given individual, as well as better identify strategies for language re-training \cite{kohn1985picture,wilshire2002aphasic,kittredge2008effect}. Early interactive activation models (see Dell; Schwartz \& Dell \cite{dell1997lexical,schwartz2006case}) captured aphasic paraphasias with single-layer architectures, but struggled to explain cross-modal error blends and latency patterns \cite{dell1997lexical}. Subsequent two-step and production-monitor approaches improved descriptive coverage yet still lacked an explicit representation of the dual semantic-phonological topology that psycholinguistic and neuroimaging work now supports \cite{foygel2000models,schwartz2006case}. Multiplex network science, with its formal treatment of inter-layer diffusion and superdiffusion, offers a principled way to model how weakened semantic-phonological coupling prolongs lexical retrieval and skews the error mix. SpreadPy operationalises this framework, allowing us to map classical aphasia phenomena onto analytically tractable network parameters.

\renewcommand\baselinestretch{1}\selectfont
\begin{table}[t!]
    \centering
    \caption{\textit{Aphasia}: Cohen's d, Starting from Phonological and Semantic Layers; Bold values if Kruskal-Wallis values are statistically significant, p<0.05; Column L stands for "Target Layer"; $\alpha_m$ stands for Max Activation, and $t_m$ stands for Time Max Activation; Legend in Errors: S: Semantic; F: formal; M: Mixed; Legend in L: Ph: Phonological, S: Semantic. }
    \label{tab:cohen_d_combined_aphasia}

    \renewcommand{\arraystretch}{1.3} 
    \setlength{\tabcolsep}{2.5pt} 
    
    \begin{tabular}{l|l|l|ccc|ccc|ccc|ccc}
        \toprule
        \textbf{Dx} & \textbf{Groups} & \textbf{L} & \multicolumn{6}{c|}{\textbf{Starting: Phonological Layer}} & \multicolumn{6}{c}{\textbf{Starting: Semantic Layer}} \\
        \hline
        & & & \multicolumn{3}{c}{\textbf{$\alpha_m$}} & \multicolumn{3}{|c}{\textbf{$t_m$}} & \multicolumn{3}{|c}{\textbf{$\alpha_m$}} & \multicolumn{3}{|c}{\textbf{$t_m$}}  \\
        \midrule
        
        & & & \textbf{R=0.2} & \textbf{0.5} & \textbf{0.8} & \textbf{R=0.2} & \textbf{0.5} & \textbf{0.8} & \textbf{R=0.2} & \textbf{0.5} & \textbf{0.8} & \textbf{R=0.2} & \textbf{0.5} & \textbf{0.8} \\
        \rowcolor{orange!10} 0.1 & S vs F & Ph & 0.01 & 0.00 & 0.03 & \textbf{0.18} & \textbf{0.11} & \textbf{0.26} & 0.29 & 0.26 & 0.11 & \textbf{0.65} & \textbf{0.58} & \textbf{0.38} \\
        \rowcolor{orange!10} 0.1 & S vs M & Ph & \textbf{0.03} & \textbf{0.02} & 0.02 & \textbf{0.40} & \textbf{0.32} & \textbf{0.38} & 0.23 & 0.23 & 0.13 & \textbf{0.48} & \textbf{0.51} & \textbf{0.41} \\
        \rowcolor{orange!10} 0.1 & F vs M & Ph & 0.02 & 0.01 & 0.01 & \textbf{0.25} & \textbf{0.23} & \textbf{0.14} & 0.08 & 0.03 & 0.02 & \textbf{0.15} & 0.07 & 0.03 \\
        \hline
        \rowcolor{orange!10} 0.1 & S vs F & S & 0.31 & 0.30 & 0.16 & \textbf{0.76} & \textbf{0.72} & \textbf{0.55} & \textbf{0.11} & 0.01 & \textbf{0.07} & \textbf{0.90} & \textbf{1.13} & \textbf{1.47} \\
        \rowcolor{orange!10} 0.1 & S vs M & S & 0.26 & 0.28 & 0.17 & \textbf{0.61} & \textbf{0.64} & \textbf{0.55} & \textbf{0.10} & 0.00 & \textbf{0.08} & \textbf{0.93} & \textbf{1.17} & \textbf{1.46} \\
        \rowcolor{orange!10} 0.1 & F vs M & S & 0.06 & 0.03 & 0.01 & 0.12 & 0.07 & 0.01 & 0.01 & 0.01 & 0.01 & 0.00 & 0.01 & 0.03 \\
        \rowcolor{purple!10} 1 & S vs F & Ph & 0.08 & 0.09 & 0.00 & \textbf{0.37} & \textbf{0.41} & \textbf{0.09} & 0.20 & 0.20 & 0.08 & \textbf{0.42} & \textbf{0.49} & \textbf{0.37} \\
        \rowcolor{purple!10} 1 & S vs M & Ph & \textbf{0.09} & 0.10 & 0.01 & \textbf{0.31} & \textbf{0.35} & \textbf{0.05} & 0.23 & 0.22 & 0.10 & \textbf{0.40} & \textbf{0.48} & \textbf{0.37} \\
        \rowcolor{purple!10} 1 & F vs M & Ph & 0.01 & 0.01 & 0.01 & \textbf{0.09} & 0.09 & \textbf{0.05} & 0.02 & 0.02 & 0.02 & 0.03 & 0.02 & 0.00 \\
        \hline
        \rowcolor{purple!10} 1 & S vs F & S & 0.18 & 0.23 & 0.12 & \textbf{0.65} & \textbf{0.69} & \textbf{0.57} & \textbf{0.03} & 0.07 & 0.07 & \textbf{0.72} & \textbf{0.79} & \textbf{0.93} \\
        \rowcolor{purple!10} 1 & S vs M & S & \textbf{0.23} & 0.26 & 0.15 & \textbf{0.70} & \textbf{0.70} & \textbf{0.57} & \textbf{0.05} & \textbf{0.09} & 0.09 & \textbf{0.77} & \textbf{0.82} & \textbf{0.95} \\
        \rowcolor{purple!10} 1 & F vs M & S & 0.04 & 0.02 & 0.02 & 0.03 & 0.00 & 0.01 & 0.01 & 0.01 & 0.02 & 0.04 & 0.02 & 0.05 \\
        \rowcolor{blue!10} 10 & S vs F & Ph & 0.01 & 0.06 & 0.03 & \textbf{0.39} & \textbf{0.52} & \textbf{0.14} & 0.12 & 0.14 & 0.04 & \textbf{0.31} & \textbf{0.40} & \textbf{0.05} \\
        \rowcolor{blue!10} 10 & S vs M & Ph & \textbf{0.03} & 0.07 & 0.01 & \textbf{0.42} & \textbf{0.49} & \textbf{0.19} & 0.18 & 0.16 & 0.01 & \textbf{0.27} & \textbf{0.35} & \textbf{0.09} \\
        \rowcolor{blue!10} 10 & F vs M & Ph & 0.02 & 0.01 & 0.02 & 0.02 & 0.03 & 0.05 & 0.05 & 0.01 & 0.03 & \textbf{0.05} & 0.06 & \textbf{0.04} \\
        \hline
        \rowcolor{blue!10} 10 & S vs F & S & 0.00 & 0.18 & 0.02 & \textbf{0.57} & \textbf{0.69} & \textbf{0.43} & 0.02 & 0.07 & \textbf{0.02} & \textbf{0.47} & \textbf{0.57} & \textbf{0.31} \\
        \rowcolor{blue!10} 10 & S vs M & S & \textbf{0.04} & 0.20 & \textbf{0.06} & \textbf{0.64} & \textbf{0.69} & \textbf{0.49} & \textbf{0.05} & 0.09 & 0.01 & \textbf{0.50} & \textbf{0.56} & \textbf{0.38} \\
        \rowcolor{blue!10} 10 & F vs M & S & 0.03 & 0.01 & 0.03 & 0.06 & 0.00 & 0.08 & 0.02 & 0.01 & 0.02 & 0.01 & 0.01 & 0.07 \\
        \bottomrule
    \end{tabular}
\end{table}
\renewcommand\baselinestretch{1.5}\selectfont

\renewcommand\baselinestretch{1}\selectfont
\begin{table}[t!]
    \centering
    \caption{\textit{Aphasia}, multiplex: Kendall Correlation between Time Max Activation ($t_m$) and Real Frequency Error (all values statistically significant)}
    \label{tab:kendall_corr_aphasia_combined}

    \renewcommand{\arraystretch}{1.4} 
    \setlength{\tabcolsep}{3pt} 
    
    \begin{tabular}{l|l|c|ccc|ccc|ccc}
        \toprule
        \textbf{Dx} & \textbf{Err.} & \textbf{Layer} & \multicolumn{3}{c|}{\textbf{Starting Phon}} & \multicolumn{3}{c|}{\textbf{Starting Sem}} & \multicolumn{3}{c}{\textbf{Starting Both}} \\
        \midrule
        & & & R=0.2 & 0.5 & 0.8 & R=0.2 & 0.5 & 0.8 & R=0.2 & 0.5 & 0.8 \\
        \rowcolor{orange!10} 0.1 & F & Ph & -0.18 & -0.18 & -0.20 & -0.17 & -0.20 & -0.20 & -0.18 & -0.19 & -0.20 \\
        \rowcolor{orange!10} 0.1 & M & Ph & -0.08 & -0.15 & -0.18 & -0.18 & -0.19 & -0.19 & -0.15 & -0.18 & -0.17 \\
        \rowcolor{orange!10} 0.1 & S & Ph & -0.04 & -0.13 & -0.21 & -0.22 & -0.25 & -0.26 & -0.15 & -0.22 & -0.25 \\
        \hline
        \rowcolor{orange!10} 0.1 & F & S & -0.14 & -0.16 & -0.18 & -0.04 & -0.09 & -0.15 & -0.13 & -0.15 & -0.19 \\
        \rowcolor{orange!10} 0.1 & M & S & -0.19 & -0.20 & -0.20 & -0.19 & -0.19 & -0.19 & -0.18 & -0.19 & -0.19 \\
        \rowcolor{orange!10} 0.1 & S & S & -0.26 & -0.25 & -0.24 & -0.25 & -0.25 & -0.24 & -0.24 & -0.24 & -0.24 \\
        \rowcolor{purple!10} 1 & F & Ph & -0.20 & -0.20 & -0.20 & -0.20 & -0.20 & -0.20 & -0.20 & -0.20 & -0.20 \\
        \rowcolor{purple!10} 1 & M & Ph & -0.17 & -0.18 & -0.18 & -0.18 & -0.19 & -0.19 & -0.18 & -0.19 & -0.19 \\
        \rowcolor{purple!10} 1 & S & Ph & -0.23 & -0.24 & -0.26 & -0.23 & -0.24 & -0.25 & -0.23 & -0.24 & -0.26 \\
        \hline
        \rowcolor{purple!10} 1 & F & S & -0.18 & -0.19 & -0.19 & -0.17 & -0.18 & -0.19 & -0.17 & -0.18 & -0.19 \\
        \rowcolor{purple!10} 1 & M & S & -0.20 & -0.21 & -0.20 & -0.20 & -0.20 & -0.21 & -0.21 & -0.20 & -0.21 \\
        \rowcolor{purple!10} 1 & S & S & -0.25 & -0.25 & -0.26 & -0.25 & -0.25 & -0.25 & -0.25 & -0.25 & -0.26 \\
        \rowcolor{blue!10} 10 & F & Ph & -0.19 & -0.19 & -0.17 & -0.20 & -0.20 & -0.17 & -0.20 & -0.20 & -0.17 \\
        \rowcolor{blue!10} 10 & M & Ph & -0.19 & -0.19 & -0.18 & -0.18 & -0.18 & -0.19 & -0.19 & -0.19 & -0.18 \\
        \rowcolor{blue!10} 10 & S & Ph & -0.23 & -0.24 & -0.23 & -0.22 & -0.24 & -0.22 & -0.22 & -0.24 & -0.22 \\
        \hline
        \rowcolor{blue!10} 10 & F & S & -0.17 & -0.19 & -0.18 & -0.18 & -0.19 & -0.18 & -0.17 & -0.19 & -0.18 \\
        \rowcolor{blue!10} 10 & M & S & -0.21 & -0.20 & -0.18 & -0.19 & -0.20 & -0.19 & -0.21 & -0.21 & -0.18 \\
        \rowcolor{blue!10} 10 & S & S & -0.24 & -0.25 & -0.26 & -0.24 & -0.26 & -0.26 & -0.24 & -0.26 & -0.26 \\
        \bottomrule
    \end{tabular}
\end{table}
\renewcommand\baselinestretch{1.5}\selectfont

\textbf{Data and modelling setup.} We combined the free association network from the Small World of Words project (semantic layer) with a phonological similarity network constructed by single-phoneme edit distance. We focused on nodes that are connected on both layers, i.e. we focused on the Largest Viable Cluster of this semantic/phonological multiplex network. This yielded a two-layer network of $N=4118$ shared word-nodes. A trial-level error corpus from the Philadelphia Naming Task project provided picture naming data from chronic anomic speakers ($N \approx 1200$ trials; equalised across error classes), used here as behavioural ground truth. For each stimulus we seeded 1.0 energy on the target word either in the semantic or phonological layer and ran deterministic diffusion for 50 time-steps under three coupling regimes ($D_x=0.1, 1, 10$) and three retention levels ($R=0.2, 0.5, 0.8$). Outcome measures were maximum activation ($\alpha_m$) and time-to-maximum activation ($t_m$) in the target layer; group contrasts were quantified with Cohen’s \textit{d} and evaluated using Kruskal-Wallis tests.

\textbf{Single-layer versus multiplex diffusion when distinguishing errors.} When differentiating semantic, formal and mixed mistakes, activation spreading on the multiplex structure (see Table \ref{tab:cohen_d_combined_aphasia}) and on single layers (see Table \ref{tab:cohen_single_layer}) provide different results. The strongest differences in $t_m$ are found when activation spreads on the semantic layer only ($d\approx 1.94, p <0.05$), followed by diffusion involving both multiplex layers but starting from the semantic one ($d\approx 1.47, p <0.05$). These differences persist across all tested retention rates and are relative to distinguishing semantic and formal (phonological) mistakes. Considerably smaller differences are found on the phonological layer alone or on the multiplex network when starting from the phonological layer ($d\approx 0.4, p <0.05$). These robust findings indicate that semantic and phonological mistakes might be differentiated by temporal dynamics more strongly related to the semantic layer only than to a mix of phonological similarities and free associations. 

\textbf{Single-layer versus multiplex diffusion for error frequency.} In terms of predicting the frequency of error types based on their diffusion dynamics, the multiplex network provides only marginally higher correlations ($\tau \approx -0.26$, see Table \ref{tab:kendall_corr_aphasia_combined}) compared to the semantic layer only ($\tau \approx -0.25$, see Table \ref{tab:kendall_corr}) for semantic mistakes. This finding confirms that the semantic layer determines most of the diffusion dynamics determining activation patterns related to semantic mistakes. Interestingly, free associations encode also activation patterns relative to mixed mistakes, since $t_m$ on both the semantic layer only and on the multiplex network exhibit similar correlations ($\tau \approx -0.20$) with the frequency of mixed mistakes. This similarity might be due to free associations being a multidimensional set of semantic but also mixed associations, as reported in prior works \cite{baker2023multiplex}. Where multiplex diffusion provides significantly stronger results than single-layer diffusion is for the correlations with the frequency of formal mistakes. These correlations are at most $\tau \approx -0.12$ in the phonological layer only, while they increase to $\tau \approx -0.20$ in the multiplex case. These patterns interestingly indicate that formal mistakes might be qualitatively different from semantic mistakes. Whereas semantic mistakes tend to occur with frequencies correlating mostly to time delays contained mostly in the semantic layer, formal mistakes occur with frequencies correlating more strongly with time delays exhibited by a multiplex phonological/semantic multiplex structure. Hence, formal mistakes might be due to an interference between semantic and phonological similarities, as posited also by past relevant works in aphasia research \cite{dell1997lexical,baker2023multiplex,dell1986spreading}.

\textbf{Single-layer versus multiplex diffusion.} The above findings indicate that single-layer diffusion never outmatches multiplex diffusion in terms of distinctions between error classes or correlations with error frequencies. Instead, multiplex diffusion can provide significant information about formal mistakes, information that cannot be retrieved in single-layer diffusion. For these reasons, in the remained of this Section, we focus on multiplex diffusion.

\textbf{Temporal dynamics - not peak magnitude - best discriminate aphasic error classes.} With reference to Table \ref{tab:cohen_d_combined_aphasia}, across all diffusion‐coupling conditions ($D_x = 0.1, 1, 10$) Cohen’s d for time-to-maximum activation is consistently an order of larger magnitude than for maximum activation. When activation is seeded in the phonological layer and measured on the same layer, $t_m$ reaches medium effect sizes (up to $d \approx 0.40$) whereas $\alpha_m$ seldom exceeds $d = 0.03$. The pattern is even clearer in the semantic layer: with semantic seeding, between semantic vs. formal and semantic vs. mixed errors, $t_m$ climbs from $d \approx 0.90$ at weak coupling ($D_x = 0.1$) to $d \approx 0.57$ at strong coupling ($D_x = 10$), while the corresponding $\alpha_m$ contrasts remain $\leq 0.11$. Hence, the latency with which a concept reaches peak activation carries the diagnostic signal, not the height of that peak.

\textbf{Semantic errors diverge earliest and most strongly.} For $D_x = 0.1$, differences in $t_m$ between semantic and formal errors in the semantic layer reach $d = 1.47$ when retention is high ($R = 0.8$), representing a very strong effect. Even at moderate retention ($R = 0.5$) the effect remains large ($d \approx 1.13$). No other error contrast approaches this magnitude, indicating that semantic-level competition is the primary driver of delayed naming in anomic aphasia.

\textbf{Superdiffusion modulates, but does not erase, temporal separation.} Increasing the coupling from $D_x = 0.1 \rightarrow 1$ halves the largest $t_m$ effect sizes (e.g. d from $1.47 \rightarrow 0.93$ for semantic vs. formal), and a further ten-fold increase ($D_x = 10$) reduces them to the upper-medium range ($d \approx 0.57$). The persistence of sizeable $t_m$ gaps even under strong coupling, i.e. in the superdiffusion regime, suggests that lexical-phonological shortcuts cannot fully compensate for degraded semantic access.

\textbf{Robust inverse link between activation time and error prevalence.} With reference to Table \ref{tab:kendall_corr_aphasia_combined}, across every simulated configuration the Kendall $\tau$ between $t_m$ and the empirical frequency of naming errors is negative ($\tau \approx -0.08, ..., -0.26$). Thus, items whose activation peaks earlier within the network are more likely to elicit a paraphasia in patients. This finding is consistent with interactive models in which quickly rising but mis-targeted activation out-competes the correct lemma before speech monitoring can intervene. Within the semantic layer $\tau$ reaches the largest magnitudes for semantic errors, ($-0.24 \leq \tau \leq -0.26$), followed by formal ($-0.14 \leq \tau \leq  -0.20$) and mixed errors ($-0.18 \leq \tau \leq -0.20$). The gradient mirrors clinical incidence rates (semantic > mixed > formal) and indicates that SpreadPy's diffusion measures on the current layers are most diagnostic for the frequency of semantic selection errors. Increasing $R$ incrementally strengthens the inverse correlation. By contrast, being in multiplex or superdiffusion regime has negligible impact on $\tau$. Whether activation is injected into the phonological, semantic or both layers simultaneously, $\tau$ estimates differ by $\leq 0.03$ within each condition. This suggests a lack of dependency on the initial multiplex or single-layer seeding. 

\textbf{Discussion of results.} Within every $D_x$ condition, larger retention values systematically inflate $t_m$ effect sizes, particularly in the semantic layer. High retention therefore prolongs intra-node accumulation, stretching the time axis and enhancing sensitivity to processing delays characteristic of different error types. Importantly, our numerical experiments indicate that, when modelling picture naming in people with aphasia, researchers should prioritise $t_m$-based metrics and explore low-to-moderate inter-layer coupling with higher retention. These settings maximise the model’s ability to reproduce the empirically observed separation between semantic, formal and mixed errors. A lack of variation in the correlations with error frequencies indicates that superdiffusion, and hence cross-layer shortcuts, matter less than temporal dispersion within layers. The stable, negative $t_m$-frequency association validates $t_m$ as a mechanistic proxy for error susceptibility. It further supports the hypothesis that aphasic naming failures are driven by prematurely peaking competitors rather than by excessively strong ones. Because the effect is parameter-robust, $t_m$ can serve as a reliable optimization target when adapting SpreadPy to individual patients or exploring therapeutic interventions that aim to prolong lexical competition in damaged semantic networks.

\section{Discussion}
\label{sec:disc}

The present work puts the spreading activation model proposed by Collins and Loftus \cite{collins1975spreading} into computational form encompassing Python and multiplex cognitive networks: SpreadPy\footnote{\url{https://github.com/dsalvaz/SpreadPy}}. Collins and Loftus' spreading activation model describes how concepts in memory are activated by energy spreading across links between conceptual representations. SpreadPy's central advantage is its ability to simulate spreading activation within both single-layer and multiplex cognitive networks \cite{stella2024cognitive}. To our knowledge, this represents a first pioneering attempt to implement spreading activation across multiple interconnected layers. This is relevant because multiplex cognitive network allow for the representation of different types of information such as semantic and phonological relationships, that are represented on distinct but interacting layers that can influence the activation dynamics of each other \cite{stella2017multiplex,stella2018multiplex,stella2024cognitive}. This enables researchers to simulate how different types of information, such as meaning and sound, interact during cognitive processes like word associations, creativity, or error production. SpreadPy builds upon the functionality of the existing \textit{spreadr} tool, implemented in R \cite{siew2019spreadr}. Our Python-based framework replicates all features of \textit{spreadr} for single-layer networks, but significantly extends the tool by supporting multiplex architectures. To our knowledge, there is currently no equivalent open-source Python tool that enables spreading activation simulations on single-layer and multiplex networks.


SpreadPy supports modelling networks in multiple languages, as demonstrated through simulations using English or Italian lexical networks. This allows researchers to work directly with data in the original language, avoiding the need for translation, which is a common limitation in many modelling tools. Modelling cognitive data in the original language is crucial, as linguistic structures such as word associations or semantic distances vary significantly across languages. Thus, translations may fail to preserve key cognitive or structural patterns specific to a certain language \cite{vigliocco2006language}.

\subsection{Technical simulation scenarios}

As a computational framework, the SpreadPy library is flexible enough to adapt to several theories and several different parameters. For instance, spreading activation can be used in networks of document similarities \cite{crestani1997application} or in cognitive networks \cite{collins1975spreading, doczi2019overview} as extensively tested in the case studies of this paper.

SpreadPy enables researchers to simulate a wide range of cognitive phenomena by flexibly adjusting key parameters in line with theoretical assumptions. For instance, varying the retention parameter $R$ allows to model how sustained activation might influence lexical retrieval. For example, in modelling the tip-of-the tongue phenomenon, low retention may simulate a temporary failure to access a given word \cite{burke1991tip}. In aphasia, different retention values may reflect impairments in maintaining activation on target nodes \cite{dell1997lexical}.

Adjusting the diffusion parameter $D_x$ \cite{gomez2013diffusion} allows control over the strength of interlayer coupling. By doing so, researchers can define whether activation spreads more readily within or across layers. This could be used for investigating mixed errors in aphasia \cite{dell1997lexical}, where interaction between layers is essential. High coupling simulates strong interactions between layers, while low coupling reflects more isolated layer functioning.

Choosing the starting point of activation allows modelling task-dependent differences. For instance, tasks involving category verification \cite{mulligan2008attention} may start from the semantic layer as they focus on meaning, while phonological similarity judgements \cite{siew2023phonological} may begin from the phonological layer due to their reliance on phonological information. Experiments involving both semantic and phonological information at the same time, such as auditory stimuli during a lexical decision task, can be modelled with activation starting from both layers simultaneously.

Another key modelling choice lies in selecting which layer/s the network is built from. Researchers focusing on phonology can simulate activation in a phonological single-layer network. This approach is well suited for experiments investigating phonological similarity judgements \cite{siew2023phonological}, spoken-word recognition \cite{vitevitch2021phonotactics} or processing spoken nonwords \cite{vitevitch1997phonotactics}. By isolating the phonological layer, SpreadPy allows users to examine effects of phonological similarity or neighbourhood density without interference from semantic associations \cite{siew2023phonological,siew2013community}. To model phonological impairments, researchers can lower the retention parameter to simulate rapid decay of activation on the phonological layer, similar to degraded phonological working memory \cite{grivol2011phonological}.  

Likewise, researchers interested in meaning-based processing can model activation spreading in semantic single-layer networks. This is relevant for investigating semantic priming effects \cite{de1983range}, category verification \cite{mulligan2008attention}, or the role of semantic distance in judgements of semantic relatedness \cite{kenett2017semantic}. Moreover, SpreadPy allows researchers to add further information as weight to nodes or links. For instance, nodes can be weighted according to their frequency which can help in explaining strong or weak priming effects \cite{yap2009individual}. 

SpreadPy's support for multiplex cognitive networks \cite{stella2024cognitive} means that experimenters can simulate scenarios in which phonological and semantic layers interact dynamically. This is particularly relevant for studies of lexical decision \cite{coltheart1979phonological}, speech-in-noise recognition \cite{bradlow2007semantic} or anomic aphasia \cite{baker2023multiplex}, where both types of information have been found to influence performance \cite{coltheart1979phonological, collins1992phonological, castro2020contributions, baker2023multiplex}. 

Beyond cognitive science, SpreadPy is also suitable for applications in information retrieval \cite{crestani1997application}. For instance, computer scientists can simulate semantic search processes in digital libraries, where nodes represent documents and edges reflect shared features such as authorship, genre, or keyword overlap \cite{vsevcech2014user}. In such contexts, spreading activation models can help evaluate how relevant documents are retrieved based on initial query activation, and how these signals evolve over time.

\subsection{Case Study Insights}

We demonstrated the utility of SpreadPy through three distinct case studies investigating spreading activation in relation to: (i) math anxiety, (ii) creativity, and (iii) aphasia.

In the affective cognition case study, SpreadPy was used to model the semantic associations of participants with varying levels of math anxiety. We used psychometric scores to classify participants into high- and low-math-anxiety groups. On group-level single-layer networks constructed from free association data, we activated the node “mathematics” and measured peak activation $\alpha_m$ and time of peak activation $t_m$ on the nodes “anxiety”, “stress” and “depression”. 

The simulations revealed that in high math-anxious individuals, “mathematics” triggered rapid activation of the concept “anxiety,” reflecting direct and strong links in the semantic-affective network. In contrast, individuals without math anxiety showed more diffuse activation paths, involving intermediate nodes that “softened” the link between math and negative affect. These findings highlight a mirroring between the structure of memory, a sustained activation of negative affect, and individual differences in math anxiety. Individuals that were classified as having high math anxiety through psychometric scores tend to maintain stronger activation of negative concepts, and their cognitive networks reflect this negative bias \cite{stella2022network}. SpreadPy thus allows researchers to operationalize how affective features are encoded in the structure of cognitive networks and to explore how these structures shape activation flow over time.

In the creativity case study, we investigated if SpreadPy could differentiate between items with varying difficulty in the Remote Associates Test (RAT). We constructed multiplex networks consisting of a semantic and a phonological layer. On either the semantic, the phonological or both layers simultaneously, we activated all three cue words (e.g. “cottage - Swiss - cake”) at the same time and measured peak activation and time of peak activation at the solution node (e.g. “cheese”). SpreadPy could effectively differentiate items by difficulty, predicting faster and higher activation peaks for easier items (see Figure \ref{fig:creativity}. Higher activation could reach solution nodes more quickly at shorter distances from the initially activated nodes. This interpretation can explain why easier RAT solutions exhibited lower $t_m$ (time of max activation) and higher $\alpha_m$ (peak activation). In contrast, more difficult items typically involve more remote associations, according to Mednick’s Associative Theory of Creativity, which proposes that creative thinking involves bridging distant semantic concepts \cite{mednick1962associative}. Our SpreadPy simulations provide a quantitative match to this theory: items rated as more difficult required more time and exhibited lower peak activation compared to easy items.

Beyond supporting established creativity theories, this case study also highlights the advantages of modelling cognitive search mechanisms relevant for creativity on a multiplex network. Our results showed that differences in RAT item difficulty were most pronounced when activation diffused through a multiplex structure combining semantic and phonological layers (see Table \ref{tab:cohen_d_creativity}). Although the semantic layer alone produced significant distinctions, reflecting the task’s inherently semantic nature, results from the multiplex network showed greater differentiation between the difficulty levels. In contrast, diffusion on the phonological layer alone produced no significant results, further underscoring that the RAT task relies mostly on semantic memory relationships rather than sound-based cues \cite{bowden2003normative,lee2014measure}. However, the combined semantic-phonological multiplex network provided additional discriminative strength, suggesting that while semantic pathways drive solution access, phonological connections may further support retrieval or strengthen activation of the solution indirectly. 

This finding supports broader psycholinguistic claims that semantic and phonological layers are tightly coupled \cite{levelt2001spoken, coltheart1979phonological, collins1992phonological,doczi2019overview} and that even semantically dominated tasks may benefit from models that can incorporate phonological information and allow for its interaction with semantic information.

The aphasia case study demonstrated how SpreadPy can be applied in clinical settings, e.g. to simulate picture-naming errors in individuals affected by disorders from the aphasic spectrum \cite{baker2023multiplex}. We constructed multiplex networks with a semantic layer and a phonological layer. We started activation at the target word (name of the object in the picture, e.g. “bottle”) and measured how quickly ($t_m$) and how much maximum activation ($\alpha_m$) arrived at the mistaken solution (e.g. “waterfall”). 

We investigated 3 different aphasic error types: semantic, formal and mixed errors. Semantic errors involve the production of words that are conceptually related to the target (e.g. saying “supporter” for “table”) \cite{kittredge2008effect}, while formal errors are driven by phonological similarity (e.g. saying “funny” for “bunny”) \cite{wilshire2002aphasic}. Mixed errors combine both semantic and phonological overlap (e.g. saying “rat” for “cat”) \cite{baker2023multiplex}. Our simulations indicate that the temporal feature of activation spreading, i.e. the time to reach peak activation, was better at discriminating between error types compared to maximum activation itself. Semantic errors were best captured using the semantic single-layer network, with clear distinctions from other error types based on $t_m$. 

This suggests a crucial advancement for aphasia studies. Our SpreadPy simulations show quantitatively that semantic mistakes arise primarily from semantic representations, the multiplex network performed almost as well in identifying semantic errors as the semantic layer alone. In contrast, formal errors were poorly differentiated from other error types in the phonological single-layer network. However, in the multiplex network, formal mistakes were far more distinguishable, suggesting that semantic information interferes with phonological retrieval. Even though formal errors may ostensibly appear to be a by-product of processes that occur on the phonological layer alone, our simulations suggest a more nuanced mechanism by which formal errors emerge from the interaction between semantic and phonological processes, rather than simply being confined to the phonological layer \cite{dell1997lexical,castro2019multiplex}. This makes the multiplex network especially valuable: it models how semantic errors arise within the semantic layer, and also captures the interference of semantic activation on the phonological layer, which helps explain formal errors.

\subsection{Investigation of superdiffusion across experiments}

Overall, our case studies on creativity and aphasia highlight the advantage of multiplex diffusion over single-layer diffusion in capturing the complex dynamics of cognitive processes. Since SpreadPy supports multiplex diffusion, it can also model superdiffusion. 
For example, in future studies on lexical retrieval, superdiffusion could explain how individuals quickly access words by drawing on both semantic and phonological associations. A word like "cat" may activate not only related concepts (e.g., "dog", "pet") but also phonologically similar words (e.g., "mat", "hat"), with cross-layer interactions amplifying the spread of activation. This multiplex advantage aligns with empirical findings of faster lexical access in healthy individuals compared to those with cognitive impairments, where layer decoupling can hinder superdiffusion \cite{stella2018multiplex,stella2024cognitive}.

Superdiffusion also has implications for learning and memory. In educational settings, students with well-integrated knowledge networks (e.g., strong connections between mathematical concepts and real-world applications) may exhibit superdiffusive activation, facilitating problem-solving learning. Conversely, conditions like math anxiety, which fragment associative networks, could disrupt superdiffusion, leading to slower and less efficient cognitive processing \cite{stella2022network,artemenko2015neural}. By modelling these dynamics, SpreadPy enables researchers to quantify the impact of network structure on cognitive performance, offering insights into individual differences and potential interventions.

Finally, superdiffusion provides a mechanistic framework for studying higher-order cognition, such as creativity and reasoning. In creative tasks, for example, the ability to combine distant concepts may rely on superdiffusive activation across semantic and contextual layers. SpreadPy's multiplex simulations allow researchers to test these hypotheses, exploring how network topology and layer coupling influence cognitive flexibility. By bridging network science and cognitive psychology, superdiffusion models can open new avenues for understanding the neural and computational principles underlying human thought.  

\subsection{Impact for Future Research}
Our findings not only demonstrate SpreadPy’s capacity for modelling cognitive dynamics in multiplex systems, but also point toward several promising directions for methodological development and future research.

From a practical standpoint, SpreadPy addresses several limitations of classical cognitive modelling frameworks. Models like Dell's interactive activation of word production \cite{dell1986spreading,dell1991mediated} are restricted by small-scale lexicons, while McClelland and Rumelhart's interactive activation model of context effects \cite{mcclelland1981interactive} rely on seemingly arbitrary parameter settings that hinder generalisation and reproducibility. Meanwhile, Collins and Loftus' \cite{collins1975spreading} spreading activation theory remains primarily verbal and difficult to operationalise. SpreadPy offers a clear computational advantage by supporting simulations on realistically sized networks with transparent and interpretable parameters and flexible parameter tuning. 

While SpreadPy is designed as a computational tool for simulating cognitive activation dynamics, its output is only as robust as the network it is given. Because the tool depends on the structural quality of the input networks, whether constructed from human association data or Large Language Models, users must pay attention to the validity and representativeness of their input networks. SpreadPy does not prescribe a single architecture or theory, and this openness is a strength, but this places responsibility on the researcher to construct transparent and valid networks. 

Notably, the case studies presented in this paper make use of only a subset of SpreadPy’s functionality. Several built-in features remain available for future exploration. For example, SpreadPy supports the integration of node- or edge-level attributes such as valence or word frequency. Future work could modulate activation dynamics using such attributes, aligning simulations more closely with known patterns of language use and word recognition. Moreover, although the current case studies employed a uniform diffusion across the network, the SpreadPy tool is able to simulate weighted spreading based on these attributes. Furthermore, inhibitory mechanisms are also implemented in the tool. We inherited the suppression parameter from Siew's \textit{spreadr} \cite{siew2019spreadr}, which globally reduces activation below a given threshold. Future refinements of this feature could simulate inhibition of specific nodes or pathways, as seen in lexical competition or error correction.

Furthermore, in contrast to many traditional cognitive models that provide static outcomes, SpreadPy allows for the tracking of activation over time. Thus, it may be combined with techniques such as EEG, eye-tracking, and mouse-tracking which emphasise the temporal unfolding of cognitive processes rather than static end states \cite{spivey2005continuous}. In addition to tracking peak activation and latency of peak activation, SpreadPy allows activation trajectories to be computed and visualised across time points and for each individual node. This enables a far more detailed comparison between empirical data and simulated dynamics than traditional models that only predict endpoint behaviour, such as reaction times. 

Finally, SpreadPy opens new opportunities for cross-linguistic simulations. The tool supports the simulation on networks in any language, making it well suited to test predictions, e.g. word recognition tasks across languages with different writing systems (e.g., alphabetic vs. logographic). For instance, researchers could compare how orthography–phonology–semantic interactions differ between English and Chinese, testing extensions of the Seidenberg and McClelland framework \cite{seidenberg1989distributed, yang2006triangle} within a multiplex network environment.

\subsection{Conclusion}

Finally, SpreadPy serves as an accessible and fully open-source tool for simulating how minds activate, associate and sometimes err through the structure of their mental networks. It facilitates theory testing, and supports empirical designs that go beyond static outcomes to capture the dynamics of cognition as it unfolds.

\section*{Acknowledgements}

\noindent This work is supported by SoBigData.it which receives funding from the European Union – NextGenerationEU – National Recovery and Resilience Plan (Piano Nazionale di Ripresa e Resilienza, PNRR) – Project: “SoBigData.it – Strengthening the Italian RI for Social Mining and Big Data Analytics” – Prot. IR0000013 – Avviso n. 3264 del 28/12/2021.  

We also acknowledge Giuseppe Fonte for useful preliminary feedback on the creativity case study dataset.

\section{Declarations}

\subsection{Funding}
We received no funding for this research.

\subsection{Conflicts of interest/Competing interests} 
We declare no conflicts of interest.

\subsection{Ethics approval} 
Ethics approval was not required as no new data were collected. All case studies are based on previously published, anonymized datasets.

\subsection{Consent to participate} 
Not applicable. No new participants were recruited as part of this study.

\subsection{Consent for publication} 
Not applicable. All data used were from publicly available sources with prior publication.

\subsection{Availability of data and materials} 
All datasets used in the case studies are publicly available through the original publications. 
For case study 1 on math anxiety, we constructed the networks based on data from \cite{ciringione2024math}.
For the multiplex networks used in case study 2 and 3, we constructed the semantic layer from the Small World of Words (SWoW) dataset \cite{de2019small}. The phonological layer was constructed from phonological similarities collected by \cite{vitevitch2008can}.
For case study 2 on creativity, we used RAT items reported in \cite{bowers1990intuition,bowden2003normative,lee2014measure,wu2020systematic}. 
Finally, the database for case study 3 on aphasic errors was used from \cite{castro2019multiplex}.

\subsection{Code availability} 
The SpreadPy library and all code used for the simulations are openly available at the following link: \url{https://github.com/dsalvaz/SpreadPy}.

\bibliographystyle{plainnat} 
\bibliography{bibliography}

\newpage

\appendix

\section{Additional results of simulations with spreading activation}

\begin{figure}[h]
\centering
\includegraphics[scale=0.72]{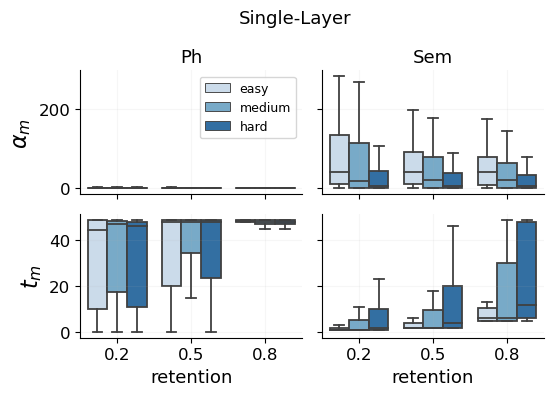}
\caption{\textit{Creativity}: Boxplots displaying maximum activation ($\alpha_m$) and time to reach it ($t_m$) at different retention ($R=0.2, 0.5, 0.8$) rates seeding diffusion only on the phonological (Ph) or the semantic (Sem) layer.}
\label{fig:creativity_single}
\end{figure}

\begin{table}[h]
    \centering
    \caption{\textit{Creativity, single-layer}: Cohen's d values for a single-layer diffusion. Bold values indicate statistically significant Kruskal-Wallis results (p<0.05).}
    \label{tab:creativity_single_layer}

    \begin{tabular}{llccc|ccc}
        \toprule
        \textbf{Single-Layer} & \textbf{Group Comparison} & \multicolumn{3}{c|}{$\alpha_m$} & \multicolumn{3}{c}{$t_m$} \\
        \midrule
        & & \textbf{R=0.2} & \textbf{R=0.5} & \textbf{R=0.8} & \textbf{R=0.2} & \textbf{R=0.5} & \textbf{R=0.8} \\
        Ph & easy vs hard & 0.17 & 0.13 & 0.10 & 0.02 & 0.06 & 0.05 \\
        Ph & easy vs medium & 0.13 & 0.15 & 0.15 & 0.11 & 0.15 & 0.05 \\
        Ph & medium vs hard & 0.02 & 0.03 & 0.07 & 0.13 & 0.10 & 0.00 \\
        Sem & easy vs hard & \textbf{0.58} & \textbf{0.63} & \textbf{0.67} & \textbf{0.44} & \textbf{0.56} & \textbf{0.60} \\
        Sem & easy vs medium & 0.10 & 0.18 & 0.23 & 0.19 & 0.28 & 0.31 \\
        Sem & medium vs hard & \textbf{0.46} & \textbf{0.43} & \textbf{0.42} & 0.30 & \textbf{0.35} & \textbf{0.30} \\
        \bottomrule
    \end{tabular}
\end{table}

\begin{table}[h]
    \centering
    \caption{\textit{Aphasia}: Cohen's d, Starting from Both Layers; Bold values if Kruskal-Wallis values are statistically significant, p<0.05; Legend: S: Semantic; F: Phonological; M: Mixed}
    \label{tab:cohen_d_both_aphasia}
    \rowcolors{2}{gray!15}{white} 

    \begin{tabular}{l|l|ccc|ccc|c}
        \toprule
        \textbf{Dx} & \textbf{Groups} & \multicolumn{3}{c|}{$\alpha_m$} & \multicolumn{3}{c|}{$t_m$} & \textbf{Layer} \\
        \midrule
        & & \textbf{R=0.2} & \textbf{R=0.5} & \textbf{R=0.8} & \textbf{R=0.2} & \textbf{R=0.5} & \textbf{R=0.8} & \\
        \rowcolor{orange!10} 0.1 & S vs F & \textbf{0.09} & \textbf{0.09} & 0.04 & \textbf{0.24} & \textbf{0.28} & \textbf{0.09} & Ph \\
        \rowcolor{orange!10} 0.1 & S vs M & \textbf{0.09} & \textbf{0.10} & 0.06 & \textbf{0.06} & \textbf{0.19} & \textbf{0.04} & Ph \\
        \rowcolor{orange!10} 0.1 & F vs M & 0.01 & 0.01 & 0.02 & \textbf{0.19} & \textbf{0.10} & \textbf{0.06} & Ph \\
        \hline
        \rowcolor{orange!10} 0.1 & S vs F & \textbf{0.07} & 0.00 & \textbf{0.03} & \textbf{0.87} & \textbf{0.92} & \textbf{1.08} & S \\
        \rowcolor{orange!10} 0.1 & S vs M & \textbf{0.06} & 0.01 & \textbf{0.04} & \textbf{0.87} & \textbf{0.94} & \textbf{1.08} & S \\
        \rowcolor{orange!10} 0.1 & F vs M & 0.01 & 0.01 & 0.01 & 0.03 & 0.01 & 0.02 & S \\
        \rowcolor{purple!10} 1 & S vs F & 0.07 & 0.08 & 0.00 & \textbf{0.39} & \textbf{0.45} & \textbf{0.22} & Ph \\
        \rowcolor{purple!10} 1 & S vs M & 0.10 & 0.10 & 0.03 & \textbf{0.36} & \textbf{0.41} & \textbf{0.20} & Ph \\
        \rowcolor{purple!10} 1 & F vs M & 0.03 & 0.02 & 0.02 & 0.05 & 0.06 & 0.02 & Ph \\
        \hline
        \rowcolor{purple!10} 1 & S vs F & \textbf{0.05} & 0.01 & 0.01 & \textbf{0.68} & \textbf{0.73} & \textbf{0.74} & S \\
        \rowcolor{purple!10} 1 & S vs M & \textbf{0.02} & \textbf{0.03} & \textbf{0.03} & \textbf{0.74} & \textbf{0.76} & \textbf{0.76} & S \\
        \rowcolor{purple!10} 1 & F vs M & 0.02 & 0.02 & 0.02 & 0.04 & 0.02 & 0.04 & S \\
        \rowcolor{blue!10} 10 & S vs F & 0.01 & 0.04 & \textbf{0.06} & \textbf{0.36} & \textbf{0.46} & \textbf{0.09} & Ph \\
        \rowcolor{blue!10} 10 & S vs M & 0.05 & 0.06 & \textbf{0.03} & \textbf{0.35} & \textbf{0.43} & \textbf{0.14} & Ph \\
        \rowcolor{blue!10} 10 & F vs M & 0.04 & 0.02 & 0.03 & 0.02 & 0.04 & 0.05 & Ph \\
        \hline
        \rowcolor{blue!10} 10 & S vs F & \textbf{0.07} & 0.02 & \textbf{0.04} & \textbf{0.52} & \textbf{0.62} & \textbf{0.37} & S \\
        \rowcolor{blue!10} 10 & S vs M & \textbf{0.04} & 0.04 & 0.01 & \textbf{0.58} & \textbf{0.63} & \textbf{0.43} & S \\
        \rowcolor{blue!10} 10 & F vs M & 0.02 & 0.02 & 0.03 & 0.05 & 0.00 & 0.07 & S \\
        \bottomrule
    \end{tabular}
\end{table}

\begin{table}[h]
    \centering
    \caption{\textit{Aphasia, single-layer}: Cohen's d, phonological (Ph) or semantic (S) layers alone; Bold values if Kruskal-Wallis values are statistically significant, p<0.05; $\alpha_m$ stands for Max Activation, and $t_m$ stands for Time Max Activation; Legend in Errors: S: Semantic; F: formal; M: Mixed; Legend in L: Ph: Phonological, S: Semantic. }
    \label{tab:cohen_single_layer}
    
    \begin{tabular}{llccc|ccc}
        \toprule
        \textbf{Single-Layer} & \textbf{Group Comparison} & \multicolumn{3}{c|}{$\alpha_m$} & \multicolumn{3}{c}{$t_m$} \\
        \midrule
        & & \textbf{R=0.2} & \textbf{R=0.5} & \textbf{R=0.8} & \textbf{R=0.2} & \textbf{R=0.5} & \textbf{R=0.8} \\
        Ph & S vs F & \textbf{0.02} & 0.01 & 0.01 & 0.08 & 0.07 & 0.02 \\
        Ph & S vs M & \textbf{0.03} & \textbf{0.06} & \textbf{0.03} & \textbf{0.54} & \textbf{0.56} & \textbf{0.41} \\
        Ph & F vs M & \textbf{0.05} & \textbf{0.06} & \textbf{0.04} & \textbf{0.45} & \textbf{0.48} & \textbf{0.37} \\
        S & S vs F & \textbf{0.05} & \textbf{0.05} & \textbf{0.06} & \textbf{1.07} & \textbf{1.40} & \textbf{1.94} \\
        S & S vs M & \textbf{0.02} & \textbf{0.08} & \textbf{0.09} & \textbf{1.14} & \textbf{1.46} & \textbf{1.82} \\
        S & F vs M & 0.02 & 0.02 & 0.03 & 0.01 & 0.01 & 0.03 \\
        \bottomrule
    \end{tabular}
\end{table}

\begin{table}[h]
    \centering
    \caption{\textit{Aphasia, single-layer}: Kendall Correlation between Time Max Activation ($t_m$) and Real Frequency Error. In bold we are highlighting correlations that are statistically significant ($p < 0.05$).}
    \label{tab:kendall_corr}
    
    \begin{tabular}{l|l|ccc|c}
        \toprule
        \textbf{Error Type} & \textbf{Single-Layer} & \textbf{R=0.2} & \textbf{R=0.5} & \textbf{R=0.8} & \textbf{} \\
        \midrule
        F & Ph & \textbf{-0.11} & \textbf{-0.12} & \textbf{-0.12} & \\
        S & Ph & \textbf{0.08} & 0.06 & 0.04 & \\
        M & Ph & 0.04 & 0.02 & 0.03 & \\
        F & S & -0.03 & -0.03 & -0.04 & \\
        S & S & \textbf{-0.25} & \textbf{-0.25} & \textbf{-0.24} & \\
        M & S & \textbf{-0.21} & \textbf{-0.20} & \textbf{-0.20} & \\
        \bottomrule
    \end{tabular}
\end{table}

\end{document}